\documentclass[10pt,journal,compsoc]{IEEEtran}

\usepackage{graphicx}
\usepackage[utf8]{inputenc} % allow utf-8 input
\usepackage[T1]{fontenc}    % use 8-bit T1 fonts
\usepackage[hidelinks]{hyperref}       % hyperlinks
\usepackage{url}            % simple URL typesetting
\usepackage{booktabs}       % professional-quality tables
\usepackage{amsfonts}       % blackboard math symbols
\usepackage{nicefrac}       % compact symbols for 1/2, etc.
\usepackage{microtype}      % microtypography
\usepackage{xcolor}         % colors

\usepackage{multirow}

\usepackage{comment}
\usepackage{float}
\usepackage{bm,amsmath}
\usepackage[inline]{enumitem}
\usepackage{subcaption}
\usepackage{mathtools}
\usepackage{natbib}
\usepackage{amssymb}

\usepackage{wrapfig}

\usepackage[nocompress]{cite}

\def\Re{\mathbb{R}}

\def\Nat{{\rm I\kern\pIR N}}

\def\argmin{\mathop{\rm arg\,min}}

\newcommand{\EE}[1]{\exptE\left[#1\right]}

\newcommand{\defeq}{\overset{\text{\tiny def}}{=}}

\def\A{{\mathcal{A}}}

\def\E{{\mathcal{E}}}

\def\S{{\mathcal{S}}}

\newcommand{\States}{\S}
\newcommand{\Actions}{\A}

\def\vec0{{\boldsymbol{0}}}

\newtheorem{thm}{Theorem}[section]
\newtheorem{lem}[thm]{Lemma}

\newtheorem{prop}[thm]{Proposition}
\newcommand{\eproof}{$\null\hfill\square$}
\newenvironment{proof}{\par\noindent{\bf Proof:\ }}{\eproof} %{\hfill\BlackBox\\[.3mm]}

\newcommand{\beq}{\begin{equation}}
\newcommand{\eeq}{\end{equation}}
\newcommand{\beqa}{\begin{eqnarray}}
\newcommand{\eeqa}{\end{eqnarray}}
\newcommand{\beqan}{\begin{eqnarray*}}
\newcommand{\eeqan}{\end{eqnarray*}}
\newcommand{\ben}{\begin{eqnarray*}}
\newcommand{\een}{\end{eqnarray*}}

\def\tr{^\top\!}

\renewcommand{\EE}[2]{\mathbb{E}_{#1\!\!}\left[#2\right]}

\newcommand{\CEE}[3]{\EE{#1}{{#2}~\middle\vert~{#3}}}
\renewcommand{\CEE}[3]{\EE{#1}{{#2}\mid{#3}}}
\def\CE#1#2{\CEE{\,}{#1}{#2}}

\def\E#1{\EE{\,}{#1}}

\ifx\theorem\undefined
\newtheorem{theorem}{Theorem}
\fi

\def\sign{\text{sign}}

\newcommand{\clip}[2]{\text{clip}_{#1}\left(#2\right)}
\newcommand{\huber}[2]{p_{#1}\!\left(#2\right)}
\newcommand{\Fhuber}{{\mathcal{F}_{\text{Huber}}}}

\def\MSBE{\text{MSBE}}
\def\MSPBE{\text{MSPBE}}
\def\MABE{\text{MABE}}
\def\MHBE{\text{MHBE}}
\def\MHPBE{\text{MHPBE}}
\def\MAPBE{\text{MAPBE}}

\def\vweights{{\theta}}
\def\hweights{{\theta_{h}}}

\def\AllFuncs{\mathcal{F}_{\text{all}}}
\def\ParamFuncs{\mathcal{F}}

\def\SignedFuncs{\ParamFuncs_{\text{sign}}}
\def\ClippedFuncs{\ParamFuncs_{\text{clip}_\tau}}
\newcommand{\Vset}{{\mathcal{F}_v}}
\newcommand{\ParamFuncH}{\ParamFuncs_h}

\newcommand{\Bop}{\mathcal{T}}
\newcommand{\BOptop}{\mathcal{T}^*}
\def\ProjH{\Pi_{\ParamFuncH,d}}

\newcommand{\approxv}{{v_\theta}}
\newcommand{\approxq}{{q_\theta}}

\def\anote#1{\triangleright \text{ #1}}

\def\revision#1{{#1}}

\DeclarePairedDelimiter\abs{\lvert}{\rvert}
\DeclarePairedDelimiter\norm{\lVert}{\rVert}

\makeatletter
\let\oldabs\abs
\def\abs{\@ifstar{\oldabs}{\oldabs*}}

\let\oldnorm\norm
\def\norm{\@ifstar{\oldnorm}{\oldnorm*}}
\makeatother

\begin{document}
\title{Robust Losses for Learning Value Functions}

\author{
  Andrew~Patterson,
  Victor~Liao,
  Martha~White
}

% The paper headers
\markboth{IEEE Transactions on Pattern Analysis and Machine Intelligence, VOL. 45, NO. 5, MAY 2023}%
{Patterson \MakeLowercase{\textit{et al.}}: Robust Losses for Learning Value Functions}
% The only time the second header will appear is for the odd numbered pages
% after the title page when using the twoside option.
%
% *** Note that you probably will NOT want to include the author's ***
% *** name in the headers of peer review papers.                   ***
% You can use \ifCLASSOPTIONpeerreview for conditional compilation here if
% you desire.

\IEEEtitleabstractindextext{
\begin{abstract}
Most value function learning algorithms in reinforcement learning are based on the mean squared (projected) Bellman error.
However, squared errors are known to be sensitive to outliers, both skewing the solution of the objective and resulting in high-magnitude and high-variance gradients.
To control these high-magnitude updates, typical strategies in RL involve clipping gradients, clipping rewards, rescaling rewards, or clipping errors.
While these strategies appear to be related to robust losses---like the Huber loss---they are built on semi-gradient update rules which do not minimize a known loss.
In this work, we build on recent insights reformulating squared Bellman errors as a saddlepoint optimization problem and propose a saddlepoint reformulation for a Huber Bellman error and Absolute Bellman error.
We start from a formalization of robust losses, then derive sound gradient-based approaches to minimize these losses in both the online off-policy prediction and control settings.
We characterize the solutions of the robust losses, providing insight into the problem settings where the robust losses define notably better solutions than the mean squared Bellman error.
Finally, we show that the resulting gradient-based algorithms are more stable, for both prediction and control, with less  sensitivity to meta-parameters.
\end{abstract}
}

\maketitle
\IEEEdisplaynontitleabstractindextext
% \IEEEpeerreviewmaketitle

\section{Introduction}

Many algorithms in reinforcement learning (RL) are built on objectives that use squared errors.
Temporal difference (TD) learning \citep{sutton1988learning} and its many variants use a semi-gradient update based on squared TD-errors and were later shown to minimize a squared projected Bellman error \citep{antos2008learning,sutton2009fast}.
Residual gradient algorithms \citep{baird1995residual,baird1999reinforcement} use a biased gradient to minimize a mean squared Bellman error (MSBE), with later work deriving unbiased variants \citep{dai2017learning,feng2019kernel}.
% Andy - I'm not sure these sentences are related to above paragraph
% The Bellman error corresponds to the difference in the approximated value of a state and the expected bootstrapped estimate: expected reward plus the expected value in the next state.
% Large differences are magnified by squaring.

Squared errors, however, tend to magnify incorrect predictions; encouraging the function approximator to expend limited representation resources on states and actions that induce the highest TD-error.
High magnitude TD-errors can occur during the learning process and can persist even for the best approximation due to state aliasing.
When two states are aliased, they are restricted to having (nearly) identical value estimates.
However, if these states have notably different rewards and next states, then this restriction can result in high Bellman error, even at convergence.
Squared errors overly emphasize even a small number of such states, potentially at the cost of accuracy in many other states.

This focus on large Bellman errors can be particularly undesirable in control.
As an example, consider the CliffWorld domain.
The agent starts in one corner of a grid and seeks to walk alongside a cliff to reach the opposite corner on the same wall.
If the agent steps into the cliff, it receives a high-magnitude negative reward and must start again.
The agent otherwise receives -1 reward per step until it reaches the goal and the episode terminates.
In order to successfully solve this domain, the agent need only learn that actions which step into the cliff yield more negative return than actions which step toward the goal.
Representing the exact magnitude of this expected negative return is unnecessary.
Squared errors do just the opposite: by squaring large errors, they expend their limited representation capacity focusing on those states.
As a result, the representation may suffer for other states and actions providing a suboptimal ordering over actions. Further, during the optimization, these high-magnitude errors can skew updates.

This challenge has been addressed heuristically through a variety of approaches in RL, which
include clipping rewards \citep{hessel2018multitask}, errors \citep{mnih2013playing}, and gradients \citep{vanhasselt2016deep}; careful manipulation of the reward function \citep{brockman2016openai,young2019minatar}; and variance reduction methods \citep{wang2016dueling,hessel2018rainbow}.
Some approaches, such as manipulating the reward function, require extensive domain knowledge and are not generally possible for all problem settings.
Other approaches, such as clipping, inhibit analyzing the fixed-point of the update.
Error clipping in Q-learning algorithms---often referred to as minimizing a Huber loss---is built on a semi-gradient update, which does not follow the gradient of any loss function.
This makes analysis of the fixed-point challenging, leaving open the question: what is the effect of clipping the error on TD-like algorithms?

In this work, we develop a robust loss to address the issues presented by mean squared errors: the mean Huber Bellman error (MHBE).
We first motivate the MHBE by showing several settings where the fixed-point under the MHBE is significantly better than the MSBE in terms of minimizing the value error.
We then address how to optimize the MHBE, relying on the insight that the MSBE can be reformulated into a saddlepoint problem\footnote{
  One of the reasons we chose the Huber loss is that it is convex and so is equal to its biconjugate; we require this for the reformulation and subsequent algorithms. Other robust losses, such as the Tukey biweight loss, are non-convex; it would not be clear how to optimize these losses.
} with the introduction of an auxiliary learned variable \citep{dai2017learning,dai2018sbeed}.
Using the same strategy, we derive a biconjugate form for the MHBE amenable to simple gradient-descent techniques. We then derive a more practical algorithm, that facilitates optimizing the robust objective under (limited) function approximation, resulting in a mean Huber projected BE (MHPBE). We show that, in this setting, our Huber Bellman error becomes a squared projected Bellman error, providing similar fixed-points to standard TD algorithms but notably improved optimization surfaces.
We empirically show that our algorithm 1) significantly outperforms existing deep RL algorithms that clip TD errors, and that we can do so without using target networks and 2) that the use of this Huber loss has a significant stabilizing effect during learning.

%We highlight the relationship between these statistically robust losses and current best-practice heuristics in deep RL, showing that algorithms derived from the MHBE significantly outperform existing deep RL algorithms which use a Huber function to clip TD errors, but do not minimize a Huber loss.

% Our key contributions
% - We introduce and analyze two new loss functions for RL
% - We derive several new algorithms for minimizing these loss functions
% - We empirically investigate which properties of MDPs impact the solutions of these new loss functions, something that was previously impossible without a formal definition of these robust losses.
% - We show that the newly introduced gradient TD algorithms significantly outperform existing semi-gradient methods and that they allow us to avoid target networks.

% Some other discussions of robustness:
%  * Henderson et al and reliability/reproducibility of results (robustness as a result of better empiricism)
%  * Safe-rl (robustness via exploration)
%  * Clipping (robustness via problem alteration)

% Some uses of the Huber loss in rl:
%  * Original DQN implementation to avoid big gradients
%  * Distributional RL to estimate arbitrary percentiles
%     * Sometimes because we actually want these percentiles, and sometimes because we want them for better optimization properties via auxiliary tasks.

\section{Problem Formulation}
We model the agent's interactions with its environment as a Markov Decision Process (MDP), \( (\States, \Actions, P, R, \gamma) \).
At each time-step $t$, the agent observes state $S_t \in \States$, selects an action $A_t \in \Actions$ according to policy $\pi : \States \to \Delta(\Actions)$, transitions to the next state $S_{t+1} \in \States$ according to transition function $P : \States \times \Actions \to \Delta(\States)$, and receives a scalar reward signal $R_{t+1}$ and discount $\gamma_{t+1} \in [0,1]$. The discount depends on the transition (state, action and next state), and encodes termination when $\gamma_{t+1} = 0$ \citep{white2017unifying}.

For the prediction setting, the agent's goal is to estimate the value function $v_\pi$ for a given policy. The value function can be defined recursively, using the Bellman operator
\begin{equation*}
  (\Bop v)(s) \defeq \CE{R_{t+1} + \gamma_{t+1} v(S_{t+1})}{S_t = s}
\end{equation*}
where the expectation is taken with respect to the policy $\pi$ and transition dynamics $P$.
The true values $v_\pi$ are the fixed point for this operator: $\Bop v_\pi = v_\pi$.
Our goal is to approximate $v_\pi$, with $\approxv \in \Vset$ for some (parameterized) function space $\Vset$. Typically, the quality of this approximation is evaluated under the value error, either mean squared value error (MSVE) or mean absolute value error (MAVE)
\begin{align*}
  \text{MSVE}(\approxv) &\defeq \sum_{s \in \States} d(s) \left( \approxv(s) - v_\pi(s) \right)^2 \\
  \text{MAVE}(\approxv) &\defeq \sum_{s \in \States} d(s) |\approxv(s) - v_\pi(s) |
\end{align*}
where $d: \States \rightarrow [0,1]$ is typically the visitation distribution under a behavior policy.\footnote{
  For exposition, we assume discrete states and actions throughout this paper.
  The connection to continuous state-actions is straightforward; we will highlight and address where this connection is less obvious.
}

One objective used to learn approximation $\approxv$ is the mean squared Bellman error (MSBE)
\begin{align}
  \MSBE(\vweights) &\defeq \sum_{s \in \States} d(s) ((\Bop \approxv)(s) - \approxv(s))^2 \\
  &= \sum_{s \in \States} d(s) \CE{\delta(\vweights)}{S_t=s}^2 \nonumber
\end{align}
where $\delta(\theta) \defeq R_{t+1} + \gamma_{t+1} \approxv(S_{t+1}) - \approxv(S_{t})$.
If $v_\pi \in \Vset$, then there exists a $\vweights$ such that $\approxv = v_\pi$ and $\MSBE(\vweights) = 0$. Otherwise, if $v_\pi \notin \Vset$, then this objective trades off Bellman error $\CE{\delta(\vweights)}{S_t=s}$ across states $s$.

This trade-off is impacted by the weighting $d$ as well as the fact that a squared error is used. The function approximator focuses on states with high weighting $d$; however, by using a squared error it also heavily emphasizes states with higher error, which may not be desirable.
In the next section, we develop an approach to use more robust losses in place of squared error.
The same approach as above can also be used for control to approximate the optimal action-values $q^*$. These values can similarly be defined using a Bellman optimality operator
\begin{equation*}
  (\BOptop q)(s, a) \defeq \CE{R_{t+1} \!+ \gamma_{t+1} \max_{a' \in \Actions} q(S_{t+1}, a')}{\!S_t\!=\!s,\! A_t\!=\!a}
\end{equation*}
with $\BOptop q^* = q^*$. The corresponding mean squared Bellman error for learning approximate $\approxq$ is
\begin{equation*}
 \sum_{s \in \States, a \in \Actions} d(s,a) \left[ (\BOptop\approxq)(s, a) - \approxq(s,a) \right]^2
\end{equation*}
where we overload $d$ to be a state-action weighting, which typically corresponds to the state-action visitation frequency.

\section{The Utility of Robust Bellman Errors}\label{sec:fixed-points}

In this section, we define the mean Huber BE (MHBE) and investigate properties of its minima.
We likewise define the mean absolute BE (MABE), allowing us to study the extrema of the MHBE tuning parameter which interpolates between MABE and MSBE.
These three Bellman errors have the same fixed-point in the tabular setting, but can have notably different fixed-points under function approximation.
We design several small MDPs to empirically show the differences between the solutions under these objectives.

\subsection{Conceptual Motivation for Robust Bellman Errors}

The robust Bellman errors are straightforward to specify,
\begin{align}
  \text{MABE}(\vweights) &\defeq \sum_{s\in\States} d(s) \abs{\CE{\delta(\vweights)}{S=s}} \\
  \text{MHBE}(\vweights) &\defeq \sum_{s\in\States} d(s) \huber{\tau}{\CE{\delta(\vweights)}{S=s}},
\end{align}
where $\huber{\tau}{\cdot}$ is the Huber function and is defined as
\begin{equation*}
  \huber{\tau}{a} \defeq \begin{cases}
    a^2 & \text{if } |a| \leq \tau \\
    2\tau |a| - \tau^2 & \text{otherwise}
  \end{cases}
\end{equation*}
for some $\tau \ge 0$.
Using $\tau=1.0$ corresponds to using squared error when the magnitude of the error is below 1, and absolute error otherwise.
% For both these losses, and the MSBE, the goal is to find values that approximate expected return.

To gain some intuition on how these objectives differ, consider a two-state MDP where the agent observes only a single signal between states.
The agent starts in the first state and deterministically transitions to the second state, where the agent remains with high probability or terminates the episode with low-probability.
Because the length of each episode varies greatly and because the first state is visited infrequently, the distribution of TD errors becomes skewed by large errors in the first state.
Updating the prediction to decrease the high error in the first state harms the prediction in the second state due to the state aliasing.
\revision{Because the true value function is not representable, the minimizer of each objective must trade-off prediction error in each state.}

\begin{figure}[h]
  \centering
  \includegraphics[width=0.8\columnwidth]{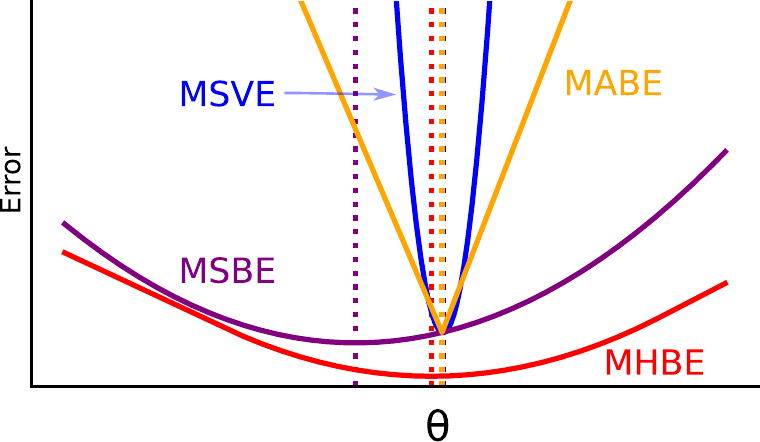}
  \caption{\label{fig:objectives}
    Objectives and fixed-points on the HardAlias-2 MDP (defined in Section~\ref{sec:prediction_experiments}).
    Dotted lines are drawn at the minima.
    The fixed-points of the robust objectives are much better proxies for the MSVE.
    Note that the MAVE (not shown) has a similar fixed-point to the MSVE, so the MHBE and MABE also well approximate the fixed-point of the MAVE.
  }
\end{figure}

Figure~\ref{fig:objectives} visualizes this trade-off among objectives.
Surprisingly, even when our goal is to minimize a mean \emph{squared} VE, the mean \emph{absolute} BE provides the closest solution among the Bellman objectives.
The MHBE provides a close approximation as well as a better optimization surface. The MSBE, on the other hand, defines a poor fixed-point, highly skewed by large errors in the first state of the MDP.
\revision{This trade-off is due to a particular property of the MDP, specifically the high skew in TD errors and the aliased states. In Section~\ref{sec:fixed-point-experiment}, we further investigate properties of MDPs for which each respective loss defines better minima.}

Much of the theory for the MSBE highlights that it provides a bound on the MSVE.
Naturally, we would like to have similar results for the MHBE to motivate that it acts as a proxy to the MAVE, as in Theorem~\ref{thm:upper-bound}.\@
It is well-known that absolute errors are hard to optimize and so the Huber error acts as a smooth approximation to the absolute error.
We use this connection to show that the MABE bounds the MAVE and the MHBE smoothly approximates the MABE.
The proof is in Appendix~\ref{app:bound_proof}.
\begin{theorem}\label{thm:upper-bound}
  For arbitrary $v \in \Re^{|\States|}$ and $0 < \epsilon < \min(\tau, 1)^2$ we have
  \begin{align*}
    &\|v_\pi - v\|_1 \leq \|(I - P_{\pi,\gamma})^{-1}\|_1 \|\Bop v - v\|_1 \\
    &\leq \|(I - P_{\pi,\gamma})^{-1}\|_1 \sum_{s=0}^d \left(\frac{\sqrt{\epsilon}}{2\epsilon} \huber{\tau}{\Bop v_s - v_s} + \frac{\sqrt{\epsilon}}{2} \right)
  \end{align*}
  \revision{
    where $P_{\pi, \gamma}$ is the substochastic transition dynamics matrix with transition-dependent $\gamma$.
  }
\end{theorem}

\revision{
  By approximating the absolute error with a Huber error, we induce an irreducible bias term $\tfrac{\sqrt{\epsilon}}{2}$ which is controlled by the Huber parameter $\tau$.
  As $\tau$ approaches zero, the Huber function approaches the absolute function and this term likewise goes to zero.
  We can observe the effect of this bias in Figure~\ref{fig:objectives} by noticing the MABE and MHBE do not share the same fixed-points.
}

\subsection{Contrast to Previous Uses of Robust Losses}

\revision{
  In supervised learning, robust losses are used to mitigate issues with high-variance, stochastic targets.
  The target for Bellman errors, however, is an expectation with no variance.
  Instead, the robust Bellman losses impact the accumulation of error across states.
  In the tabular setting, all three Bellman errors share the same solution: the true expected return. Under function approximation, they distribute errors differently.
}

More similar to their use in supervised learning, distributional RL uses methods which can learn robust statistics of the target, i.e. the return, by using a modified Bellman residual \citep{dabney2017distributional,bellemare2017distributional}.
Distributional RL methods no longer seek to learn expected returns.
Even in the tabular setting, the optimal value function learned by distributional RL methods will be different from the optimal value function defined by the robust Bellman errors studied in this work.
In fact, these two approaches are complimentary: one could seek to find the median return by minimizing the median error over states.

Huber-like losses have become commonplace in deep reinforcement learning implementations \citep{raffin2021stablebaselines3,dhariwal2017openai,mnih2013playing}.
Although these implementations often claim to be minimizing a Huber loss, the iterative weight update does not minimize any known loss function.
They apply the Huber function to the TD error, $p_\tau\left( R_{t+1} + \gamma_{t+1} v_{\theta_\text{old}}(S_{t+1}) - v_\theta(S_t)\right)$, with a fixed target network $v_{\theta_\text{old}}$.
This update more closely resembles that of a mean Huber TD error, rather than Bellman error, because the Huber is inside the expectation: $\mathbb{E}[p_\tau( \delta_t ) | S_t = s] \neq p_\tau\left(\mathbb{E}[ \delta_t  | S_t = s] \right)$.
Further, the target network causes the objective to change with time.
Analytically, the fixed-point of this DQN update remains an open question.
Empirically, clipping the magnitude of the TD errors has led toward better learning stability on certain problem settings \citep{mnih2013playing}, however this insight is inconsistent when validated across a wider testbed of problem settings \citep{obando-ceron2021revisiting}.
By instead defining a Huber BE objective, we can concretely characterize the loss surface and solutions of the proposed objective.

\subsection{Experiment: Comparing Fixed-points}\label{sec:fixed-point-experiment}

Before developing online learning algorithms to minimize the robust Bellman objectives, we first seek to empirically understand the properties of these objectives and under what circumstances they provide better or worse solutions than the squared Bellman error.
To do so, we investigate six different problem settings, each chosen to highlight particular properties of the objectives.
Our goal is not to show that any one objective necessarily dominates the others, but rather to find problem settings where each objective performs particularly well and likewise where each performs poorly.

\revision{We assume that the state is fully observable and the agent constructs some features of this state, say by a neural network or tile-coding.
For the following problems, we will assume that the feature generating function is fixed and given to the agent; for now, we only study the linear mapping from features to values.
This models a common setting where the agent receives a complete Markov state, learns a feature generating function from those states, then learns a linear value function from those features---e.g. using an end-to-end neural network learning procedure such as DQN.}
Although the agent has access to the true underlying state, we cannot assume that the feature generating function will always yield useful features which cleanly discriminate between states of highly different value, especially early in the learning process.

The first two problem settings build on simple MDPs that have state representations which aggressively alias multiple states into a single feature.
Because this aliasing skews the TD errors, we expect to find that the robust objectives perform well compared to the squared Bellman error.
\textbf{HardAlias-1} is an 8-state random walk where the first, third, and final states share a common feature, and the remaining five states share three features.
\textbf{HardAlias-2} is the 2-state problem from \citet{tsitsiklis1997analysis}, which was originally designed to highlight the insufficiency of minimizing the squared Bellman Residual, with lightly modified reward so the optimal value function cannot be perfectly represented.
% We lightly modify the reward function of the MDP so the optimal value function cannot be perfectly represented allowing each objective to have different minima.

The next investigated problem setting (\textbf{Outlier}) highlights the advantages of the MHBE by creating a single outlier state with a large magnitude return among a large set of states with approximately normally distributed returns.
We use a randomly initialized frozen neural network to generate five features with a one-hot state encoding as input.
The agent starts in a state that has an $\epsilon=0.01$ chance of terminating immediately with -1000 reward, or a $1-\epsilon$ chance of entering the middle state of a 49-state random walk.

The final two problems are chosen to highlight a scenario where the MSBE finds favorable solutions compared to the robust objectives, which we expect will behave more conservatively in this idealized setting.
In these problems, the returns are distributed approximately normally across states and states are lightly aliased.
We use two random walks, the first with $N=5$ states (\textbf{SmallChain}) and the other with $N=19$ states (\textbf{BigChain}), with a randomly initialized neural network representation of size $\tfrac{N}{2}$.
The agent receives a reward of $-1$ or $+1$ on the left and right-most states respectively.

\begin{figure}[t]
  \centering

  \includegraphics[width=0.80\columnwidth]{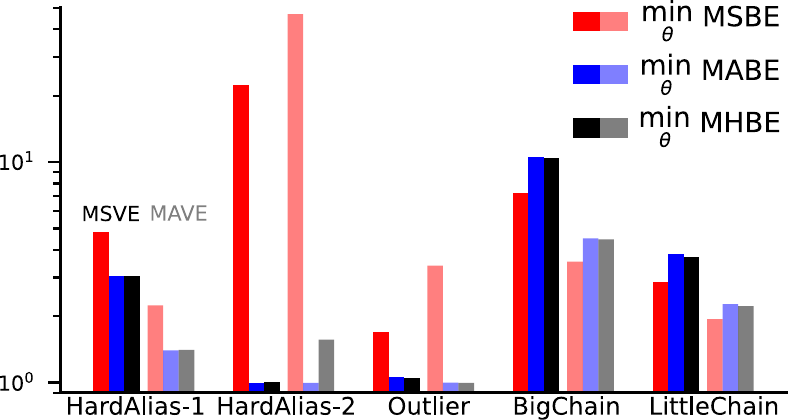}
  \caption{\label{fig:fixed-points}
    Evaluating the quality of the fixed-points of each objective function according to the MSVE and MAVE across several prediction problems.
    Error is plotted relative to the best representable value function.
    The robust losses are better in the hard aliasing domains, the MHBE is slightly better in Outlier, and the MSBE is better on the classic random walks.
  }
\end{figure}

Figure~\ref{fig:fixed-points} shows the MSVE and MAVE for each fixed-point, relative to the best realizable value function on each problem.
We find that the fixed-points of the MSBE were often poor on problems which were designed to have harmful aliasing;
in the case of HardAlias-2, the fixed-point of the squared objective was approximately twenty times worse than the MHBE.
Additionally, the robust objectives performed consistently well across all tested conditions, considerably outperforming the MSBE on the hard aliasing problems while only slightly underperforming on the idealistic chain problem settings.
Knowing nothing about our problem setting ahead of time, this might suggest one should conservatively prefer a robust objective to avoid potential catastrophic performance.

\section{Reformulating Robust Bellman Errors}\label{sec:robust_be}

In this section, we reformulate the MHBE and MABE to facilitate optimization.
We highlight a double sampling issue which, similar to the MSBE, prevents us from simply applying standard optimization routines to the Bellman objectives.
We generalize the conjugate formulation of the MSBE \citep{dai2017learning} to allow any proper, convex, and lower semi-continuous error function including the square, absolute, and Huber functions.
Finally, we show that by parameterizing these objectives, the biconjugate Bellman errors belong to a class of generalized \emph{projected} Bellman errors \citep{patterson2022generalized}. This connection between robust Bellman errors and the squared projected Bellman error underlying TD provides a new robustness perspective on the utility of projected errors.

\subsection{The Double Sampling Issue}

Though the MHBE is straightforward to specify, it is not necessarily straightforward to optimize.
The difficulty is obtaining a sample of the gradient of this objective for the same reason as the MSBE\@: the \emph{double sampling} issue.\footnote{
  Note that the use of robust losses in supervised learning does not result in the same problem. In supervised learning it is straightforward to take gradients through the robust losses. The use of Bellman operators in our objectives, with our value function approximator in the targets, is the root of these difficulties, regardless of using the squared, Huber or absolute errors. The targets in supervised learning are unbiased samples, constant with respect to the parameters of the function approximatior.
}
To see why, let us examine the gradient of the MSBE
\begin{equation*}
 \nabla \MSBE(\vweights)\! =\!\sum_{s \in \States} d(s) \CE{\delta(\vweights)}{S\!=\!s} \nabla \CE{\delta(\vweights)}{S\!=\!s}
 .
\end{equation*}
To sample this gradient requires a sample of $\delta(\vweights)$ for the first expectation and an independent sample of $\delta(\vweights)$ for the second expectation.
Otherwise, using the same sample, we estimate the gradient of $\CE{\delta(\vweights)^2}{S=s}$ instead of $\CE{\delta(\vweights)}{S=s}^2$.
Due to the chain rule and the nonlinearity of $\abs{\cdot}$ and $\huber{\tau}{\cdot}$, both the MABE and MHBE will suffer from the same issue as the MSBE.

This {double sampling} issue has made it difficult to design practical algorithms to minimize the MSBE \citep{baird1995residual,sutton2018reinforcement}.
The most naive approach---simply ignoring the issue and reusing the same sample for both expectations---results in a biased objective known as the mean squared TD error.
This objective is well-known to be a poor proxy for the value error, resulting both in slow learning algorithms and inaccurate solutions \citep{antos2008learning,sutton2018reinforcement}.

An alternative approach is to update weights using only a part of the gradient, deemed a \emph{semi-gradient}.
This semi-gradient approach is the underlying update mechanism for a wide-class of reinforcement learning algorithms, the temporal difference (TD) family of algorithms which includes its namesake algorithm, TD \citep{sutton1988learning}, as well as more modern algorithms such as DQN \citep{mnih2013playing}.
These non-gradient updates are not guaranteed to converge to any fixed-point, diverging towards infinity for even simple MDPs \citep{baird1995residual}.
Algorithms constructed on these semi-gradient updates have suffered many downstream consequences as a result, such as soft-divergence with neural network weights \citep{vanhasselt2018deep}, or requiring a suite of domain-specific heuristics to induce convergence \citep{obando-ceron2021revisiting,ghiassian2020gradient}.

More recently, the double sampling issue has been addressed using a reformulation of the MSBE using conjugate functions \citep{dai2017learning,dai2018sbeed}. The reformulation results in a saddlepoint optimization problem, where a secondary estimator is used to estimate one of the expectations in the gradient. This approach has since been used to obtain a generalized objective \citep{patterson2022generalized}, that unifies these methods for the MSBE and the gradient TD methods developed for linear function approximation, including GTD \citep{sutton2008convergent}, GTD2 and TDC \citep{sutton2009fast}, and TDRC \citep{ghiassian2020gradient}. The objective is called the mean squared \emph{projected} Bellman error (MSPBE), because the secondary estimator acts like a projection.
\revision{
  We use the same conjugate approach in the next section to define the MHBE in a way that is more amenable to sample-based optimization.
  In the following section, we then connect the MHBE to a projected error which forms the basis for our optimization algorithms.
}

\subsection{Conjugate Bellman Errors}
To facilitate optimizing the MHBE and MABE, we
reformulate the objectives using biconjugates. For a real-valued function $f: \Re \rightarrow \Re$, the conjugate is $f^*(h) \defeq \sup_{x \in \Re} xh - f(x)$. This function $f^*$ also has a conjugate, $f^{**}$, which is called the biconjugate of $f$. Further, for any function $f$ that is proper, convex, and lower semi-continuous, the biconjugate $f^{**}(x) = f(x)$ for all $x$ by the Fenchel-Moreau theorem \citep{fenchel1949conjugate,moreau1970infconvolution}.
Because the three functions we want to reformulate---the absolute, Huber, and square functions---are all proper, convex, and lower semi-continuous, this equivalence allows us to reformulate these losses using biconjugates to avoid the double sampling issue without changing the solutions to these losses or even the loss surface.

To rewrite existing results in our notation and provide some intuition, we first write the reformulation for the MSBE.
The conjugate of the squared error $f(x) = \tfrac{1}{2}x^2$ is $f^{*}(h) = \max_{x \in \Re} hx - \tfrac{1}{2} x^2$, which is in fact again the squared error: $f^{*}(h) = \tfrac{1}{2} h^2$ (this result is well known, but for completeness we include the proof in Appendix~\ref{app:biconjugates}). The biconjugate is $f^{**}(x) = \max_{h \in \Re} xh - \tfrac{1}{2} h^2$ and $f(x) = f^{**}(x)$. We can use this to get that, for $x = \CE{\delta(\vweights)}{S = s}$,
\begin{equation*}
\CE{\delta(\vweights)}{S = s}^2 = \max_{h \in \Re} 2\CE{\delta(\vweights)}{S = s}h -  h^2
.
\end{equation*}
If we have function space $\AllFuncs$---the set of all possible functions $h: \States \to \Re$---then we get that
\begin{align*}
  \MSBE(\vweights) &=  \sum_{s \in \States} d(s) \max_{h \in \Re} (2 \CE{\delta(\vweights)}{S = s}h - h^2)\\
 &= \max_{h \in \AllFuncs} \sum_{s \in \States} d(s) (2 \CE{\delta(\vweights)}{S = s}h(s) - h(s)^2)
\end{align*}
where the maximization comes out of the sum using the interchangeability property \citep{shapiro2014lectures,dai2017learning} and $h(s)$ is a function that allows us to independently pick a maximizer for every state in the summation.

If we have the maximizing function $h^*(s)$, it is straightforward to sample the gradient.
Because $h^*(s)$ itself is not directly a function of $\vweights$, then the gradient is
\begin{align*}
  \nabla_\vweights &\sum_{s \in \States} d(s) ( 2 \CE{\delta(\vweights)}{S_t = s}h^*(s) - h^*(s)^2) \\
  &= \sum_{s \in \States} 2 d(s) h^*(s) \CE{\gamma \nabla_\vweights v(S') - \nabla_\vweights v(s))}{S_t = s}.
\end{align*}
Drawing samples $S \sim d(\cdot)$, $A \sim \pi(\cdot | S)$, and $S' \sim P(\cdot | S, A)$, we can easily compute a stochastic sample of the gradient. In practice, we simply optimize the resulting saddlepoint problem with a minimization over $\vweights$ and maximization over $h$. Note the optimal $h^*(s) = \CE{\delta(\vweights)}{S_t=s}$.

We can use the same procedure for the Huber error and the absolute error. We derive the biconjugate form for the Huber error in the following proposition. Though it is a relatively straightforward result to obtain, to the best of our knowledge, it is new and so worth providing formally.
\begin{prop}\label{prop:huber}
  The biconjugate of the huber function is $f_{\tau}^{**}(x) = \max_{h \in [-\tau, \tau]} xh - \frac{1}{2} h^2$.
\end{prop}

The absolute value has biconjugate $\max_{h \in [-1, 1]} xh$.
As in the squared error case, this is a well known result but we include the proof for completeness in Appendix~\ref{app:biconjugates}.
Notice that for both the MHBE and MABE, we have a constrained optimization problem for $h$, which differs from the MSBE.

We can now provide the forms for MABE and MHBE:\@
\begin{align*}
  \MABE(\vweights) &\defeq \!\!\max_{h \in \SignedFuncs}\!\! \sum_{s \in \States} d(s) h(s) \CE{\delta(\vweights)\!}{\!S=s} \\
  \MHBE(\vweights) &\defeq  \!\!\max_{h \in \ClippedFuncs} \!\!\sum_{s \in \States} d(s) (2 h(s) \CE{\delta(\vweights)\!}{\!S=s} - h(s)^2)
\end{align*}
$\SignedFuncs$ is the set of all functions $h_{\text{sign}} \!:\! \States \!\to\! \{-1, 1\}$ and $\ClippedFuncs$ the set of all functions $h_{\text{clip}_\tau} \!:\! \States \!\to\! [-\tau, \tau]$.

\subsection{Projected Robust Bellman Errors}

In practice we will generally have parameterized functions $v$ and $h$, and so the biconjugate objectives will no longer perfectly obtain the maximum $h^*(s)$.
As a result, for parameterized $h$ and finite samples, we will only approximate the original MABE and MHBE objectives, just as we only approximate the original MSBE with the current gradient TD family of algorithms.
This approximation can actually be seen as a projection on the Bellman errors, previously highlighted for the MSBE \citep{patterson2022generalized} and which we show in this section for the MHBE.

In the finite state setting, we can represent the parameterized function $h \in \ParamFuncH$ as a vector $u \in \Re^{|\States|}$ composed of entries $\CE{\delta(S)}{S=s}$; thus the vector $u = \Bop v_\vweights - v_\vweights$.
We can define a projection operator on $u$ as
\begin{align*}
  \ProjH u \defeq \argmin_{h \in \ParamFuncH} \| u - h \|_d
\end{align*}
for convex subset $\ParamFuncH$, where $d: \States \to [0, 1]$ is a weighting over states. For the Huber, we further restrict $h$, to get $\Fhuber = \mathcal{F}_{\text{clip}} \cap \ParamFuncH$. Because $\mathcal{F}_{\text{clip}}$ is convex, $\ParamFuncH$ is convex by assumption and the intersection of convex sets is convex, we get that $\Pi_{\mathcal{F}_{\text{Huber}}, d}$ is a valid projection.

We can therefore use the same approach as \citet[Section 4.3]{patterson2022generalized}, because we simply have a slightly different convex set over the space of functions for $h$. Consequently, using exactly the same steps, we get that
\begin{align}
\max_{h \in \Fhuber}\! \sum_{s \in \States} d(s) &(2 \CE{\delta}{s}h(s) \!-\! h(s)^2) \nonumber \\
&\defeq \| \Pi_{\Fhuber, d} (\Bop v_\vweights - v_\vweights) \|^2_d.
\end{align}
This result says that the $\MHBE$ with parameterized $h$ is in fact an MSPBE!
In other words, it seems we have started with a Huber error and returned to a squared error.
\revision{
  Importantly, this squared projected Bellman error differs slightly from the typical $\MSPBE$.
  The projection is onto a different set of functions, $\mathcal{F}_{\text{clip}}$, as opposed to the set of all linear functions $\mathcal{F}$.
  This projection plays the role of deemphasizing errors of high-magnitude.
}

Using this connection to projected BEs, we revisit our discussion of fixed-points now assuming a parameterized $h$.
In Section~\ref{sec:fixed-points}, we computed the fixed-point of the MSBE assuming that the class of functions for $h$ was unconstrained, that $h \in \mathcal{F}$.
\revision{
Likewise, for the MHBE, the only imposed constraint on $h_\text{clip}$ was that it was any function with range $[-\tau, \tau]$, namely $h_\text{clip} \in \mathcal{F}_{\text{clip}}$.
}
Let us now consider projected BEs. More specifically for the MSPBE defined for TD, we have the same set for $h$ and $v$: $\ParamFuncH = \Vset$.
To define the MHPBE with $h \in \Vset$, we project the Bellman errors to $\Vset$ and further restrict them to have range $[-\tau, \tau]$; we denote this function class $\mathcal{F}_{\text{clip},v}$.
%In the linear function approximation setting, the composition of these projections $\tilde{h} = \Pi_{X_\text{clip}} h = \Pi_{\mathcal{F}_\text{clip}} \Pi_X h$ yields a linear function $\tilde{h}(s) = \hweights\tr \mathbf{x}(s)$ where $\tilde{h}(s) \in [-\tau, \tau]$.
%

Intuitively, we might expect that changing the projection operator from $\Pi_\Vset$ for the MSPBE to $\Pi_{\mathcal{F}_{\text{clip},v}}$ for the MHPBE would result in different fixed-points.
Both objectives, however, share the same fixed-point. The reason is that this TD fixed-point already has zero projected BE; further projection has no effect. We formalize this in the below theorem.
\begin{theorem}
Let $\Vset$ be a convex set of functions for $v_\vweights$, where $\MSPBE(\vweights) \defeq \| \Pi_{\Vset, d} (\Bop v_\vweights - v_\vweights) \|^2_d$. Let $\mathcal{F}_{\text{clip},v} \defeq \mathcal{F}_{\text{clip}} \cap \Vset$. Then the solution is the same for the MSPBE and the MHPBE with $\Fhuber = \mathcal{F}_{\text{clip},v}$ and $\tau > 0$. Further, this solution has zero error under both objectives.
\end{theorem}
\begin{proof}
Let $\vweights^*$ be the solution to the MSPBE. Then
\begin{align*}
\MHPBE(\vweights^*)
&= \| \Pi_{\Fhuber, d} (\Bop v_{\vweights^*} - v_{\vweights^*}) \|^2_d \\
&\le \| \Pi_{\ParamFuncs_{\text{clip}}, d} \Pi_{\Vset, d}  (\Bop v_{\vweights^*} - v_{\vweights^*}) \|^2_d \\
&\le  \| \Pi_{\Vset, d}  (\Bop v_{\vweights^*} - v_{\vweights^*}) \|^2_d  = 0
\end{align*}
The first inequality follows from the fact that projecting onto the set $\Vset$ first and then projecting further to $\mathcal{F}_{\text{clip}}$ cannot have a smaller norm than projecting directly to $\Fhuber$. The second inequality follows from the fact that the projection is a contraction under $d$, again by definition.

Further, when $\MSPBE(\vweights) > 0$ then $\MHPBE(\vweights) > 0$ for some $\vweights$.
We show this by contradiction. Assume that for some $\vweights$, $\MSPBE(\vweights) > 0$ and $\MHPBE(\vweights) = 0$.
This implies that $h(s) = 0$ for all $s \in \States$ for the MHPBE, however because $h \in \Fhuber$ and $\Fhuber \subseteq \Vset$, then $h$ must also be in $\Vset$ and so likewise $\MSPBE(\vweights) = 0$, yielding a contradiction.
% --------- nullspace argument ---------
% because the null space of $\Pi_{\Fhuber,d}$ is $(\ParamFuncs_{\text{clip}} \cap \Vset)^\perp = \{\mathbf{0}\} \cup \text{Null}(\Vset) = \text{Null}(\Vset)$ which correspondingly is the null space of $\Pi_\Vset$.
% Because the norm $\| \mathbf{u} \|_d^2 = 0$ only when $\mathbf{u} = \mathbf{0}$, and because the projections for the MSPBE and MHPBE share the same null space, then $\MHPBE(\vweights) = 0$ if and only if $\MSPBE(\vweights) = 0$.
\end{proof}
%
%To see why, recall that the TD fixed-point is $\vweights_\text{TD} = \mathbf{A}^{-1} \mathbf{b}$ for $\mathbf{A} \defeq X\tr (I - P_\gamma)\tr X$ and $\mathbf{b} \defeq X\tr R$.
%Then
%\begin{align*}
%  \text{M}&\text{HPBE}(\vweights_\text{TD}) = \|\Pi_{X_\text{clip}} \left( \Bop v - v \right) \|^2_d \\
%  &= \| \Pi_{\mathcal{F}_\text{clip}} X (X\tr X)^{-1} X\tr \left( \Bop v - v \right) \|^2_d \\
%  &= \| \Pi_{\mathcal{F}_\text{clip}} X (X\tr X)^{-1} \left( -\mathbf{A} \mathbf{A}^{-1}\mathbf{b} + \mathbf{b} \right) \|^2_d \\
%  &= \| \Pi_{\mathcal{F}_\text{clip}} X (X\tr X)^{-1} \mathbf{0} \|^2_d \\
%  &= \| \Pi_{\mathcal{F}_\text{clip}} \mathbf{0} \|^2_d = 0
%\end{align*}
%because for any choice of $\tau >= 0$, clipping the zero vector element-wise still results in the zero vector.

\begin{figure}[ht]
  \centering
  \includegraphics[width=0.75\columnwidth]{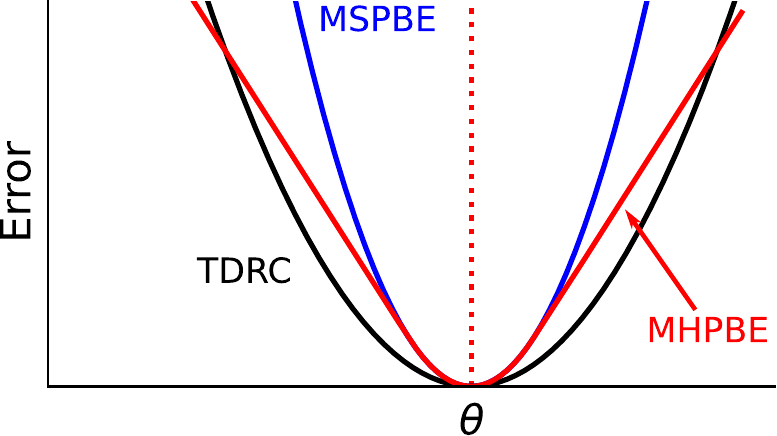}

  \caption{\label{fig:projected-losses}
    Visualizing the loss surface for the MSPBE (blue), the MHPBE (red), and an approximation of the loss followed by the TDRC algorithm (black).
    TDRC does not define a valid projection, but we can compute the $\ell_2$ regularized solution for $h$ for each $\theta$, to plot the idealized loss surface.
    The MHPBE is a squared projected loss, but the projection under the Huber flattens the surface for large residuals, characteristic of the flat regions of the Huber function.
    TDRC has a less sharp surface for a local region near the fixed-point, but ultimately suffers from using a squared loss for very large residuals.
  }
\end{figure}

This connection gives insight into why certain algorithms optimizing the MSPBE have improved stability properties.
The TDRC algorithm \citep{ghiassian2020gradient} optimizes the MSPBE, but with a regularization term on the parameters of $h(s)$.
This regularization plays a similar role to clipping, as it prevents $h$ from producing too large of values. We visualize the connection in Figure~\ref{fig:projected-losses}.

The above result is only for the case where we use $\Vset$ as the function class for $h$. More generally, the minima of the MSPBE and MHPBE are not the same for $\ParamFuncH \supset \Vset$.
For example, we could add an additional set of features to those used for $\Vset$, to get a strictly larger class of functions for $\ParamFuncH$. The projected Bellman errors under $\ParamFuncH$ are no longer zero and further projection by clipping may change the solution.

\begin{figure}
  \centering
  \includegraphics[width=\columnwidth]{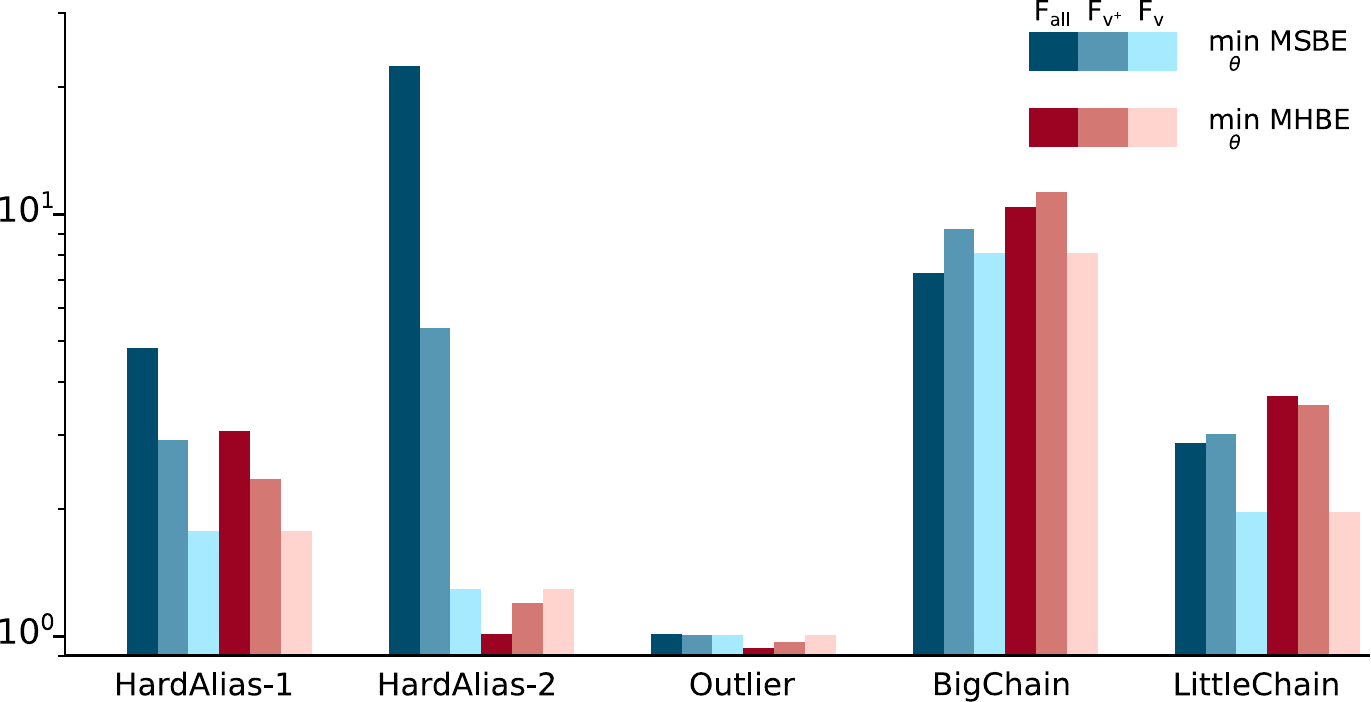}

  \caption{\label{fig:projected-fixed-points}
    Evaluating the quality of fixed-points for the projected Bellman errors with three different projection sets.
    The more saturated colors (left) correspond to no projection;
    the less saturated colors (right) correspond to using $\ParamFuncH = \Vset$.
    The interim colors represent an intermediary projection which uses five additional features to fit the Bellman residual.
  }
\end{figure}

We now revisit the fixed-points in our six environments, with these projected objectives.
In Figure~\ref{fig:projected-fixed-points}, we visualize the fixed-points for three different choices of projection for both the Huber Bellman error and the squared Bellman error.
We consider the two extremes---namely $\ParamFuncH = \mathcal{F}_{\text{all}}$ to get unprojected Bellman errors and $\ParamFuncH = \Vset$---
%, when all error not explained by $X$ is projected to zero---i.e. the projected Bellman error case---and when no error is projected away---i.e. the Bellman error case---
as well as an interim projection where $h$ uses five additional features.
To construct these five additional features, we use five random linear combinations of the original features for each problem setting then pass these combinations through a ReLU nonlinearity to simulate using a larger neural network to estimate $h$ than that used to estimate $v$.

We find that in most cases, the full projection yields the best fixed-points on these tasks. This result suggests that much of the robustness comes from projecting away excess errors.
A natural conclusion, therefore, is that the primary role of the Huber function when optimizing these projected errors will be during the transient optimization, and not in the final asymptotic solution.

\section{Optimizing the Objectives}\label{sec:optimization_algs}

In this section, we derive gradient-based updates for the MHPBE and MAPBE.
The gradient in terms of $\vweights$, for a given $h$, is actually the same: $ \sum_{s \in \States} d(s) h(s) \CE{\nabla \delta(\vweights)\!}{\!S=s}$.
This means that the job of selecting between the absolute, squared, and Huber Bellman errors rests solely on how we approximate the secondary variable, $h(s)$.

There are many ways to estimate the $h$ for the MHPBE and MAPBE.
A natural starting point would be to use the same estimate, $\tilde{h}_{\hweights}(s) \approx \CE{\delta}{S=s}$, as for the MSPBE, then apply the corresponding non-linear function to $\tilde{h}$---sign or clipping.
This gives the following updates for the objectives.
\begin{align*}
  &h(s_t) = \sign(\tilde{h}_{\hweights_{,t}}(s_t)) && \hspace{-1.5cm} \anote{$\MAPBE$} \\
  &h(s_t) = \clip{\tau}{\tilde{h}_{\hweights_{,t}}(s_t)} && \hspace{-1.5cm} \anote{$\MHPBE$} \\
  &h(s_t) = \tilde{h}_{\hweights_{,t}}(s_t) && \hspace{-1.5cm} \anote{$\MSPBE$}
\end{align*}%
\vspace{-1.9em}
\begin{align}
  \hweights_{,t+1} &= \hweights_{,t} + \alpha_h \left( \delta_t - \tilde{h}_{\hweights_{,t}}(s_t) \right) \nabla_{\hweights_{,t}} \tilde{h}_{\hweights_{,t}}(s_t) \label{eqn:h_update_gtd2} \\
  \vweights_{t+1} &= \vweights_{t} + \alpha_v h(s_t) \left( \nabla_{\vweights_{t}} v(s_t) - \gamma_{t+1} \nabla_{\vweights_{t}} v(S_{t+1})  \right) \label{eqn:v_update_gtd2}
\end{align}
Notice if we specifically parameterize $\tilde{h}_{\hweights}(s) = \hweights\tr x(s)$ and $v(s) = \vweights\tr x(s)$, then we recover the GTD2 algorithm with linear function approximation \citep{sutton2009fast}.
Because the update for the primary weights is exactly the same as GTD2 and because the clip function encodes box-constraints on the secondary weights (and so is closed and convex), convergence of the GTD2-like algorithm for the MHPBE follows directly from \citet{nemirovski2009robust}.

This parameterization may not find the best $h$ under the constraints. Another strategy is to pick a parameterized function class that encodes the constraints.
For instance, in the case of the MAPBE the function, $h(s)$ is effectively a 2-class classifier for the sign of $\CE{\delta}{S=s}$, which can be learned using logistic regression.
Likewise, for the MHPBE, the clipping function can be approximated using a rescaled logistic function $\clip{\tau}{x} \approx \tau \tanh(\tfrac{1}{\tau} x)$. In practice, we found the simpler linear approximation followed by clipping to be just as effective. We therefore pursue the strategy given in Equations \eqref{eqn:h_update_gtd2} for the rest of the paper, both because it is simpler and because it introduces fewer differences to the algorithms that optimize squared Bellman errors.
% which we will use for the nonlinear control algorithm in Section~\ref{sec:control_experiments}.
% MARTHAC: The above is also an approximation, so I don't see this as a disadvantage.
%Despite being approximations, directly parameterizing the sign and clip functions may provide some advantages such as being a smooth approximation to the sign function or a twice differentiable approximation to the clip function; improving their optimization surface.

\begin{figure*}[t]
  \centering
  % MSVE
  \includegraphics[width=\textwidth]{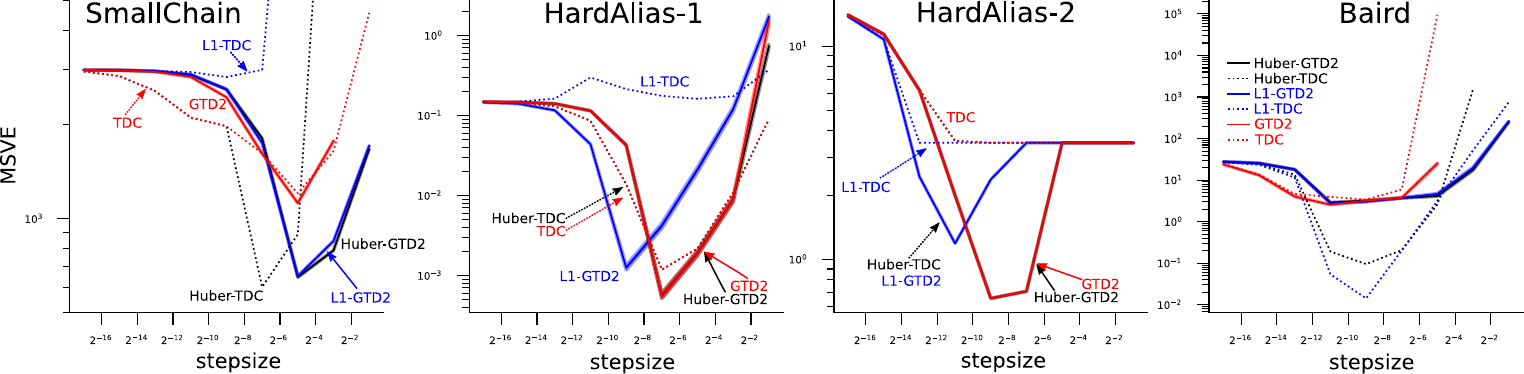}

  \caption{\label{fig:prediction}
    MSVE averaged over 100 independent trials, for each stepsize in prediction domains.
    The mean squared algorithms generally performed well across environments---even the adversarially chosen environments---suggesting the difficulty in minimizing the MABE.\@
    The Huber algorithms performed best across many environments, often displaying less sensitivity to the choice of stepsize.
  }
\end{figure*}

Several past works have reported a large performance difference between the GTD2 saddlepoint algorithm and the TDC gradient-correction algorithm \citep{white2016investigating,patterson2022generalized}.
The primary parameter vector in gradient-correction methods depends directly on a sample of the TD error, $\delta$, thus benefits from a direct unbiased error signal. The gradient correction update is
\begin{equation}\label{eq_tdc}
  \vweights_{t+1} = \vweights_{t} + \alpha_v \left( \delta(s_t) \nabla_{\vweights_{t}} v(s_t) - \gamma_{t+1} h(s_t) \nabla_{\vweights_{t}} v(S_{t+1}) \right)
\end{equation}
The saddlepoint algorithms, on the other hand, rely fully on $h$---as in Equation \eqref{eqn:v_update_gtd2}---providing a low-variance but possibly highly biased update.

Manipulating the update in Equation \eqref{eqn:v_update_gtd2} shows the relationship between the saddlepoint and gradient-correction update, for transition $s, r, s'$
\begin{align*}
  -\nabla_\vweights &\delta(\vweights) h(s) = h(s) \nabla_\vweights v(s) - \gamma h(s)  \nabla_\vweights v(s') \\
  &\hspace{-0.1cm}
  =\!(h(s) - \delta(\vweights) + \delta(\vweights))\nabla_\vweights v(s) - \gamma h(s)  \nabla_\vweights v(s') \\
  &\hspace{-0.1cm}
  =\!\underbrace{\delta(\vweights)\nabla_\vweights v(s)\!-\!\gamma h(s) \nabla_\vweights v(s')}_{\text{TDC update}}
  \!+\!\underbrace{(h(s)\!-\!\delta(\vweights)) \nabla_\vweights v(s)}_{\text{extra term}}
\end{align*}
The extra term accounts for the deviation between $\delta$ and our estimate, $h(s)$.
In fact, in the case of linear function approximation, and when $h$ is the best linear approximation, the expected value of this additional term over all states is zero \citep[Section 7.1]{patterson2022generalized}.
%making TDC an unbiased estimate of the true gradient as shown in Appendix~\ref{app:tdc_bias}.

Similarly, we can construct a corresponding gradient correction update to optimize the robust objectives, but at the cost of some additional bias in the approximate gradient because the extra term does not go away in expectation for the MHPBE or MAPBE.\@
The update for $\vweights$ stays the same, given in Equation \eqref{eq_tdc}, with our alternative update for $h$.
We report the bias of the gradient for this update, in Appendix~\ref{app:tdc_bias}, and empirically investigate both updates in Section~\ref{sec:prediction_experiments}. We can convert these equations to the off-policy setting by incorporating importance sampling ratios; we provide these modified updates in Appendix \ref{app:off-updates}.

\section{Experiment: Optimization Algorithms}\label{sec:prediction_experiments}

We have seen that the robust objectives define sensible asymptotic solutions on this set of problems; however, this leaves open the question of how well we can find these fixed-points using a stream of online, off-policy data.
% and without assuming priveledged access to the underlying MDP dynamics.
%
We investigate the effectiveness of the optimization algorithms proposed in Section~\ref{sec:optimization_algs} on a subset of the linear prediction problem settings, using one problem designed to favor the MSPBE (Small chain), two problems designed to favor the MAPBE (HardAlias 1 and 2), and Baird's counterexample.
The three objectives all have the same solutions in Baird's counterexample, because weights can be found to make the BE zero in every state. This counterexample, however, was specifically designed to highlight challenges in online optimization for temporal difference algorithms.
We would expect all optimization algorithms to perform equivalently on this problem setting, given sufficient optimization steps and appropriately decayed stepsizes; however transiently and with a fixed-stepsize, we might expect the robust objectives to have better optimization properties.

We test both the saddlepoint and gradient-correction algorithms for each of the three objectives. To respect historical naming, we label the saddlepoint algorithms with GTD2 and the gradient-correction algorithms with TDC. TDC and GTD2 are the original algorithms with a regression update for $h$. Huber-TDC and Huber-GTD2 use the same regression update for $h$, and then when using $h$ to update $\theta$, clip to range $[-\tau, \tau]$ for $\tau = 1$. L1-TDC and L1-GTD2 also learn $h$ with linear regression and then take the sign of $h$. All of the GTD2 variants use the same update to the primary weights $\vweights$, given in Equation \eqref{eqn:v_update_gtd2}. All of the TDC variants use the same update to $\vweights$, given in Equation \eqref{eq_tdc}.

 % MARTHAC: repetitive
%On each problem, we test a total of six optimization algorithms: two algorithms for each of the three objectives.
%The first optimization algorithm, called GTD2, is based on the saddlepoint formulation of each corresponding objective and provides an unbiased sample of the conjugate Bellman errors.
%This saddlepoint formulation, however, has been found to perform poorly in practice for the mean squared objective \citep{ghiassian2018online,ghiassian2020gradient}.
%The second investigated algorithm, TDC, introduces a small amount of bias into the update rule but historically has been found to have much better performance than GTD2.

Our hypothesis is that the robust objectives will provide lower variance update rules, allowing larger stepsizes and faster convergence to their fixed-points.
In order to measure this, we run each algorithm for a fixed number of steps and compute the area under the learning curve, providing a sense of both convergence rate and asymptotic error.
We look at performance across a wide range of stepsizes, $\alpha \in \{2^{-16}, 2^{-15}, \ldots, 2^{-1}\}$, and report the mean and standard error over 100 independent trials for each stepsize, algorithm, and problem setting. In later experiments, we use the same stepsize for both $h$ and $v$, to reduce the number of hyperparameters for the algorithms.
%Ultimately, our goal is to recommend an online optimization algorithm which has high performance across a diverse range of problem settings, while requiring as little hyperparameter tuning as possible.
However, for this first set of experiments, the goal is to obtain a sense of idealized performance.
%by maximizing performance over each algorithm's secondary stepsize.
We, therefore, also sweep over ratios $\eta = \frac{\alpha}{\alpha_v} \in \{2^{-4}, 2^{-3}, \ldots, 2^3, 2^4 \}$ for the stepsize $\alpha$ for $h$.

In Figure~\ref{fig:prediction} we show stepsize sensitivities for each problem and algorithm measured using the MSVE.\@
Due to the similarity in conclusions, we relegate the MAVE sensitivities to Appendix~\ref{app:additional_results}.
The L1-TDC variant appears to suffer as a result of biased gradient estimates and generally performs worse across stepsizes than its GTD2-based counterpart.
Generally the robust GTD2 algorithms and the Huber-TDC algorithm show wider stepsize sensitivities and often the MHPBE algorithms show marginally better performance for their best choice of stepsize.
On the idealized SmallChain environment, the algorithms optimizing the mean squared objective tended to be slightly less sensitive to their choice of stepsize.
However, the robust algorithms had far less error, suggesting faster convergence rates on this problem.
% MARTHAC: Maybe not true since this is the MHPBE
%  despite the robust objectives defining slightly worse fixed-points on this problem as seen in Section~\ref{sec:fixed-points}.
On the Baird counterexample problem, the robust TDC variants performed notably better than all other tested algorithms and all robust algorithms showed much less sensitivity to stepsize than the mean squared variants, suggesting overall faster convergence for the robust objectives on this problem.

% --> Distributions
\begin{figure*}[ht]
  \centering
  \includegraphics[width=0.9\textwidth]{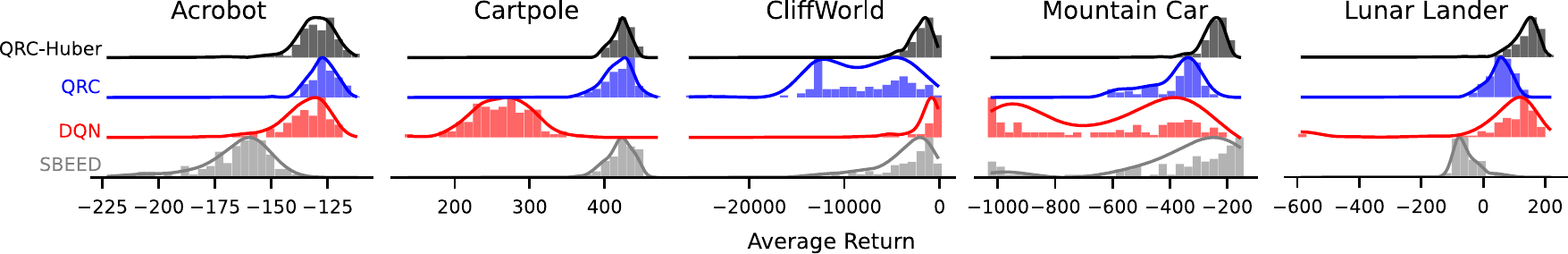}

  \caption{\label{fig:benchmark-control}
    Performance distribution over 100 random seeds \revision{for the best meta-parameter setting chosen per-algorithm and per-domain}.
    The performance measure is the average return over the last 25\% of steps.
    QRC-Huber consistently has approximately normal and narrow distributions around high-performance returns.
    DQN has inconsistent behavior, with bimodal performance on Mountain Car and Lunar Lander, and  long-tailed performance on Acrobot and Cartpole.
    SBEED has inconsistent performance with high-variance on several domains and long-tailed performance on CliffWorld and Mountain Car.
  }
  \vspace{0.5cm}
\end{figure*}
% --> Learning curves
\begin{figure*}
  \centering
  \includegraphics[width=0.9\textwidth]{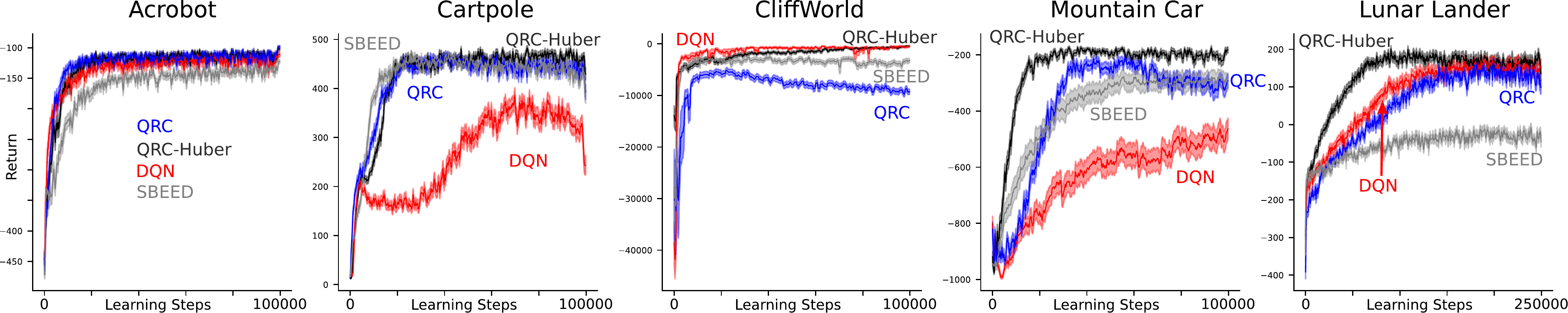}

  \caption{\label{fig:learning-curves}
    Learning curves for the best meta-parameter configuration for each domain, averaged over 100 random seeds.
    Shaded regions indicate one standard error.
    QRC-Huber is the only algorithm which is consistently among the best performing algorithms for every environment.
    DQN exhibits notable instability in both the Cartpole and Mountain Car environments, while QRC suffers from its squared loss in the adversarially designed CliffWorld environment.
    The SBEED algorithm consistently performs suboptimally on every domain except Cartpole, with notably worse performance on Lunar Lander.
  }
\end{figure*}

\section{The QRC-Huber Algorithm}\label{sec:qrc-huber}

The extension to the control setting is straightforward, due to the development of the QRC algorithm (Q-learning with Regularized Corrections) for the MSPBE \citep{patterson2022generalized}. The MSPBE can be generalized to control for a fixed state weighting, by 1) learning action-values $\approxq$ and 2) using Bellman errors based on a maximum over actions in the next state. The reformulation of the MSPBE and resulting algorithm with two parameters is otherwise the same. The only difference from QRC is in our secondary estimator $h$, which additionally incorporates clipping.

To learn $q$ and $h$, we use a two-headed neural network.
The first head estimates $\approxq(s, a)$ and the second head estimates $\tilde{h}(s, a)$. Each of these heads has one output for every action.
We block gradients from being passed back from the second head of the network, allowing the network's full function approximation resources to be used for predicting $\approxq(s, a)$ as accurately as possible.
The update rules are
\begin{align*}
&\delta = r + \gamma \max_{a'} \approxq(s', a') - \approxq(s, a)\\
  &\vweights_{t+1} = \vweights_{t} + \alpha ( \delta \nabla_\vweights \approxq(s, a) - \gamma h(s, a) \nabla_\vweights \max_{a'} \approxq(s', a'))\\
  &\hweights_{,t+1} = \hweights_{,t} + \alpha \left( \delta - \tilde{h}(s, a) \right) \nabla_{\hweights_{,t}} \tilde{h}(s, a) - \alpha\beta \hweights_{,t} \\
  &h(s,a) = \clip{\tau}{\tilde{h}(s,a)}
%  &h(s,a) = \tau \tanh(\tfrac{1}{\tau} \tilde{h}(s,a))
\end{align*}
where $\theta$ refers to all of the parameters of the neural network, except the parameters  $\hweights_{,t+1}$ for the secondary head, and $\beta$ is the regularization parameter.
Following the recommendation in the original paper \citep{ghiassian2020gradient}, we chose to keep $\beta=1$ for all experiments. This regularized version of our algorithm incorporates both robustness using clipping and regularization.

This algorithm uses a gradient-correction update. As before, we could also consider a saddlepoint update. However, we found that this had much worse performance in control, as we show in Appendix~\ref{app:additional_results}. For this reason, we only outline the gradient-correction version of our control algorithm with the Huber loss, as the proposed control algorithm for this work, and use this version for the remaining experiments.

%This parameterization was used for an algorithm called QRC, an extension of the TDRC algorithm to control \citep{ghiassian2020gradient}. As discussed in \citet{ghiassian2020gradient}---and reconfirmed in our own experiments in Appendix~\ref{app:additional_results}---using the saddlepoint update rule leads to poor performance in control, so we choose to use the gradient correction update.
%Unlike in the prediction setting, we choose to use a twice differentiable approximation of the clipping function to allow easier optimization with pseudo-second order methods like ADAM \citep{kingma2015adam}.

\section{Experiment: nonlinear control}\label{sec:control_experiments}

In this section we empirically investigate the QRC-Huber algorithm. We first compare to several control algorithms in five classic control domains, investigating both learning speeds and stability in performance over different runs. We then show that QRC-Huber performs better \emph{without} target networks, highlighting that these gradient algorithms may provide an alternative path to stabilizing DQN without slowing down learning. Finally, we conclude with a demonstration in a larger environment called Minatar.

% MARTHAC: Repetitive and doesn't give the topic paragraph for this section
%Our experiments so far focused on prediction with linear function approximation.
%However, one of the primary motivating factors of using conjugate Bellman errors is the natural and theoretically sound extension to nonlinear function approximation and control.
%In this section, we empirically investigate a Huber algorithm for nonlinear control, where we estimate $\approxq$ and $h$ using neural networks.

\subsection{Experiments in Classic Control Domains}
For the nonlinear control experiments, we investigate three classic control problems---Mountain Car, Cart-pole, and Acrobot---from the Gym suite \citep{brockman2016openai}, a larger domain with a heavily shaped reward---Lunar Lander---and one additional domain designed to be particularly challenging for squared error algorithms, Cliff World.
For all domains, we use discount factor $\gamma=0.99$ and $\epsilon=0.1$ for the $\epsilon$-greedy policy.
The episode is cut off if the agent fails to reach a terminal state in a pre-specified number of steps.
When cut off, the agent is teleported back to the start state and does not update its value function, thus preventing the agent from bootstrapping over the teleportation transition.

We compare the QRC-Huber algorithm with three baseline algorithms from prior work.
Because QRC-Huber builds on the QRC algorithm, we use this as a baseline to determine the impact of using a Huber Bellman error in place of a squared Bellman error.
All design decisions and other elements of the update rule are the same between QRC and QRC-Huber.
We additionally compare to SBEED \citep{dai2018sbeed}, the original algorithm stemming from the conjugate Bellman error work.
We modify the SBEED algorithm to use a direct mellowmax policy based on its estimated value function, instead of a parameterized policy as in its original implementation, in order to match the update rules of the other baselines.
Finally, we compare to DQN \citep{mnih2013playing}.
Due to DQN being a semi-gradient method, and because its use of a clipped error is only superficially a Huber error, we expect to find that DQN exhibits lower stability and higher sensitivity to meta-parameters.

We sweep a broad range of stepsizes and report results for every swept stepsize in Appendix~\ref{app:additional_results}.
For QRC-Huber, we fix the Huber threshold parameter $\tau=1$ for all domains except Mountain Car, where we use $\tau=2$.
We further ablate the impact of this decision in Appendix~\ref{app:additional_results}.
For the QRC methods, we chose not to use target networks---a frozen, infrequently updated set of weights for the bootstrapping target---so that we can highlight the stability provided by using true gradient-based methods with robust losses.
DQN and SBEED use targets networks and sweep over multiple refresh rates.
DQN additionally sweeps over its Huber clipping parameter and SBEED sweeps over its objective tuning parameter $\eta$ and mellowmax parameter $\lambda$.
In total, QRC and QRC-Huber tune over 6 meta-parameter combinations while DQN tunes over 120 and SBEED tunes over 360.

To demonstrate the stability of each algorithm, we report the full distribution of the performance metric over 100 independent trials for the best stepsize on each domain.
We use the average return achieved over the last 25\% of steps as our performance metric.
We expect algorithms which exhibit stable performance to have a narrow, approximately normal distribution centered around higher return, whereas we expect algorithms which are unstable to have wide performance distributions or even multi-modal distributions. We also report standard learning curves, in Figure~\ref{fig:learning-curves}.

Figure~\ref{fig:benchmark-control} shows the performance distributions of each tested algorithm.
QRC-Huber exhibits narrow and approximately normal performance distributions for every domain, suggesting the stability of the algorithm over random seeds.
The QRC algorithm performs reasonably on the Acrobot and Cart-pole domains, but performs quite poorly on the Cliff World domain.
Because QRC is based on the mean squared Bellman error, the poor performance on Cliff World is exactly as expected, since this domain was chosen adversarially to highlight challenges with mean squared errors.
While DQN is based on a clipped loss function that appears similar to the mean Huber Bellman error, it does not seem to enjoy the same stability as QRC-Huber, with average performance far worse than QRC-Huber on four of five domains due to high bimodality or long-tailed performance distributions. The learning curves in Figure~\ref{fig:learning-curves} further highlight that QRC-Huber is the most robust of the three, across all five problem setting, either having comparable or notably better performance.
\revision{
  The poor performance of DQN on Mountain Car and Cartpole is counterintuitive; given a sufficient number of learning steps and a large enough delay between target network updates, DQN can perform well in these environments.
  This results in a fundamental trade-off between stable performance and sample efficiency.
  The experiment in Figure~\ref{fig:learning-curves} favors sample efficiency due to its short length.
}

\subsection{Omitting Target Networks}\label{sec:ablations}

A primary motivation for building on gradient TD methods is that they are a theoretically sound way to obtain stable, convergent TD methods.
Target networks currently lack theoretical justification, but are believed to empirically improve the stability of semi-gradient learning rules such as in DQN.
Considering the non-negligible overlap in intended use case between gradient algorithms and target networks, we test for interactions between these algorithmic tools by extending the gradient algorithms to include target networks and measuring the change in performance.
In this section, we address the question: ``could gradient-based algorithms benefit from the use of target networks?''.

% --> Target net ablation
\begin{figure*}[ht!]
  \centering
  \includegraphics[width=0.70\textwidth]{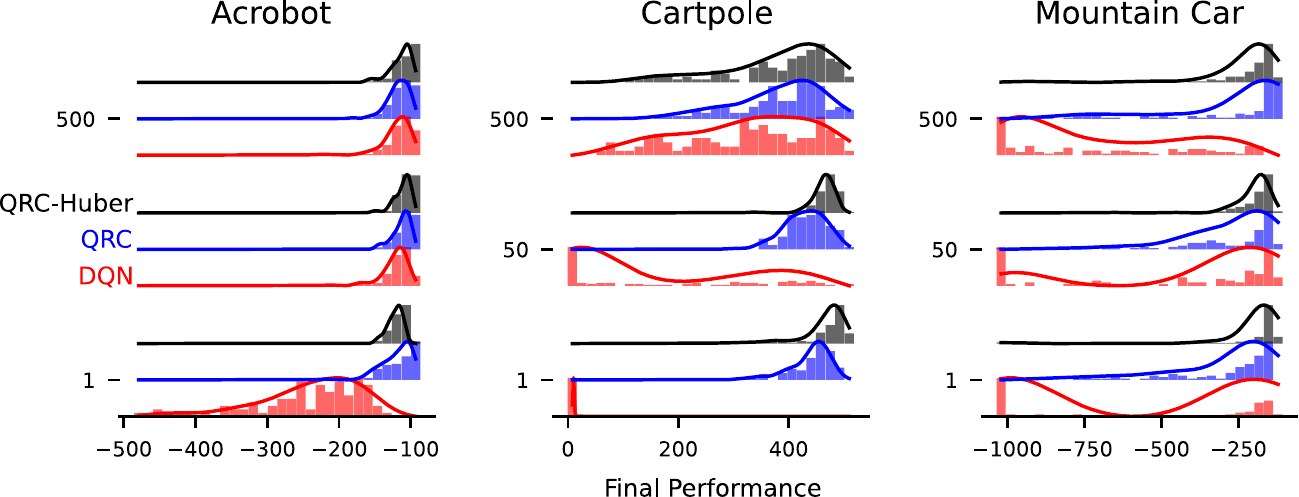}

  \caption{\label{fig:target-ablation}
      Ablating the impact of the target network refresh rate (1, 50 and 500) on the performance of the nonlinear control algorithms. A refresh rate of 1 means no target networks are used.
      DQN requires target networks to achieve above random performance on Cart-pole and to reduce the bimodality of its performance on Mountain Car.
      Even with target networks, DQN still exhibits large skew and bimodality in its performance distributions, indicating instability. The gradient methods QRC-Huber and QRC both perform better \emph{without} target networks (the last row).
  }
\end{figure*}

We introduce target networks into QRC and QRC-Huber
and sweep over the refresh rate---the number of learning steps before the target network weights are overwritten with the weights of the current network. If the target network refresh rate is one, then target networks are not used. We report the distribution of final performance over 100 random seeds, in Figure~\ref{fig:target-ablation}.
The gradient-based algorithms perform better without target networks across environments.
Because these algorithms already exhibit high stability, adding target networks serves only to harm the sample efficiency of the algorithms.
DQN always required target networks in order to solve any of the tested problems, while requiring environment-specific tuning of the target network parameter.

On the excluded CliffWorld environment, we found that all algorithms performed best without target networks. This result is unsurprising given the simplicity of the environment dynamics: CliffWorld is designed to show differences in fixed-points, rather than differences during learning.
We exclude Lunar Lander due to the high computational cost of performing extensive parameter sweeps and because QRC-Huber clearly already outperforms DQN on Lunar Lander \emph{without} target networks as shown in Figure~\ref{fig:benchmark-control}.

\subsection{Experiments on Minatar}\label{sec:minatar}

Finally, we demonstrate that QRC-Huber can scale to larger domains using more complex neural network architectures.
We use the Minatar suite of five miniaturized Atari games which retain much of the complexity of the full Atari games, while considerably reducing the computational requirements and cost \citep{young2019minatar}.
We use a convolutional neural network architecture with two hidden layers to learn value functions from images.
We demonstrate that gradient-based learning rules without target networks, such as QRC and QRC-Huber, can outperform semi-gradient learning rules such as DQN, even when DQN is allowed to additionally tune its usage of target networks.

To avoid domain overfitting and reduce the cost of meta-parameter tuning, we treat the entire Minatar suite as a single problem setting.
As such, each algorithm must pick one meta-parameter setting to use across all five games. The other benefit of this design is that it favors algorithms that are insensitive to hyperparameter choices.
We allow all three control algorithms to sweep over a small range of stepsizes and allow  DQN to additionally sweep over target network refresh rates.
We set the discount factor $\gamma=0.99$ for all domains and otherwise use the same design parameters as \citet{young2019minatar}.

In order to compare performance of each agent, we average scores over each game using probabilistic performance profiles \citep{jordan2020evaluating,barreto2010probabilistic}, then report the average scaled performance across the entire suite with 95\% confidence intervals.
We run each algorithm with its best meta-parameter setting for 30 runs on each game allowing comparisons on the Minatar suite using a total of 150 samples for each algorithm.
Additional procedural details can be found in Appendix~\ref{app:experiment_details}.

\begin{figure}[h!]
  % a little bit of magic to make figure think the label is a table, not a figure
  \makeatletter\def\@captype{table}\makeatother
  \caption{Average performance on Minatar}\label{tab:minatar}
  \centering
  \begin{tabular}{ll}
    \toprule
    QRC-Huber & $0.53 \pm 0.03$ \\
    QRC       & $0.47 \pm 0.02$ \\
    DQN       & $0.36 \pm 0.06$ \\
    \bottomrule
  \end{tabular}
\end{figure}

In Table~\ref{tab:minatar}, we report the average scaled return across games in the Minatar suite.
Despite having four times the number of meta-parameter combinations and the ability to use target networks, DQN performs considerably worse than either gradient-based algorithm.
Because QRC-Huber and QRC yield approximately normal performance distributions, we use a paired t-test and find that QRC-Huber has a statistically significant performance increase over QRC.
DQN's performance profile is notably skewed by a small number of failing runs, however the difference in performance between DQN and either gradient-based algorithm is significant according to both a paired t-test and a much less powerful bootstrap t-test (which allows for a skewed sampling distribution).
That QRC and QRC-Huber perform similarly is unsurprising as the largest possible reward in any Minatar game is $+1$, a design decision made in part because many algorithms---such as DQN---are unstable when learning from large rewards.
Additional results on the Minatar suite are included in Appendix~\ref{app:additional_results}.

\section{Conclusion}

In this work, we extended the saddlepoint reformulation of the mean squared Bellman error, introducing a novel pair of robust losses, the mean absolute Bellman error and the mean Huber Bellman error.
We demonstrated that the solutions to these robust objectives are comparable to the MSBE, and in some scenarios are significantly better according to the value error.
The resultant gradient-based algorithms are less sensitive to choice of stepsize in prediction and have more stable performance distributions in control.

We considered robust losses to learn expected returns. This modification is complementary to robust estimates of value, like median returns and percentiles. For distributional RL, for example, quantile regression and a modification using a quantile Huber loss has been used \citep{dabney2017distributional,bellemare2017distributional}. But, like clipping in DQN, this was applied to the TD update with distributional Bellman operators without accounting for the semi-gradient update. Developing these conjugate forms for this broader class of value estimation problems is an exciting next step.

% \begin{ack}
%   % TODO:
% \end{ack}

\bibliographystyle{mybibstyle}
\bibliography{paper}

\begin{IEEEbiography}[{\raisebox{0.7cm}{\includegraphics[width=1in,height=1.25in,clip,keepaspectratio]{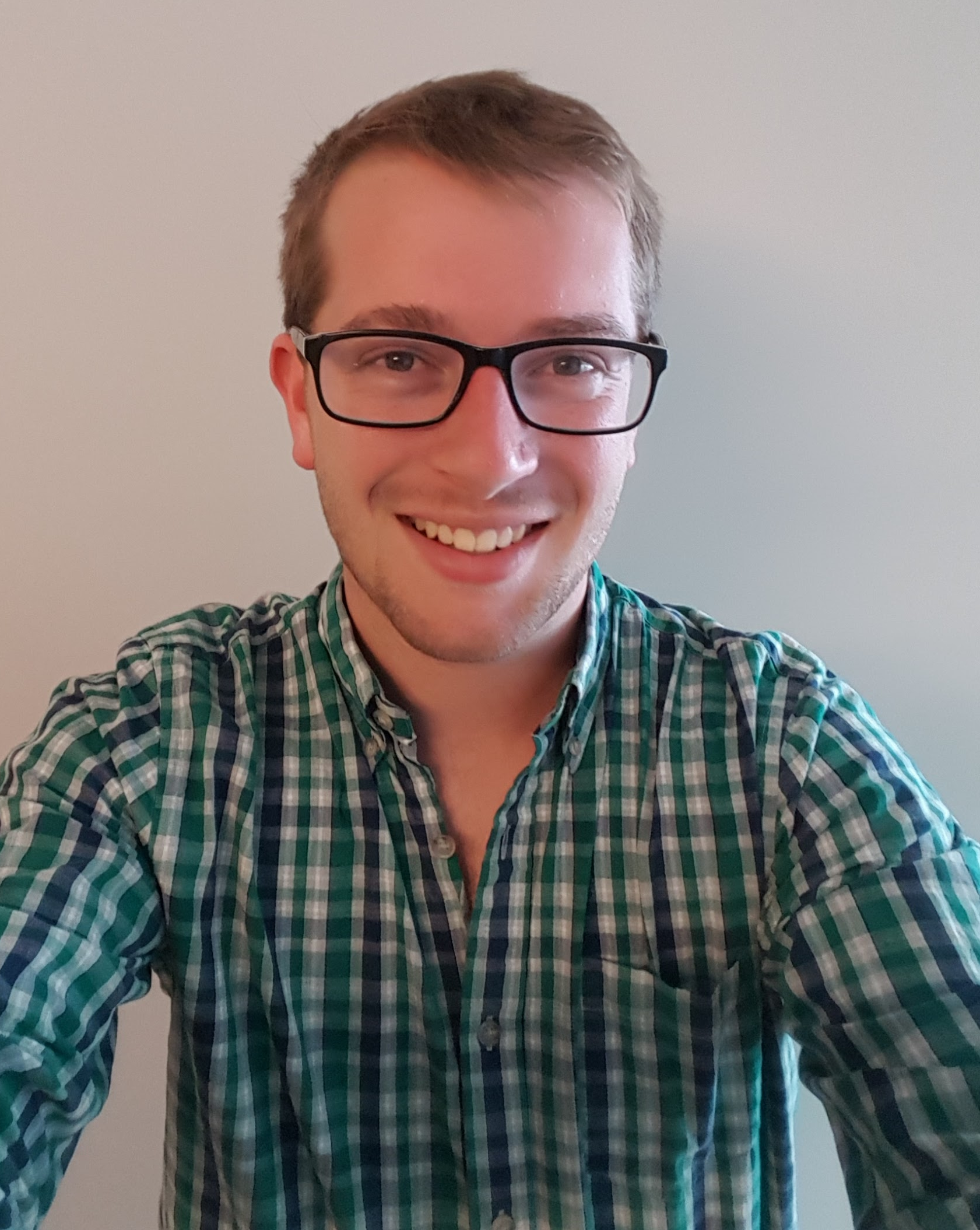}}}]{Andrew Patterson}
  is working toward a PhD in Statistical Machine Learning at the University of Alberta.
  His primary research interests include reinforcement learning, particularly in developing and understanding sound off-policy learning algorithms.
\end{IEEEbiography}

\begin{IEEEbiography}[{\raisebox{0.7cm}{\includegraphics[width=1in,height=1.25in,clip,keepaspectratio]{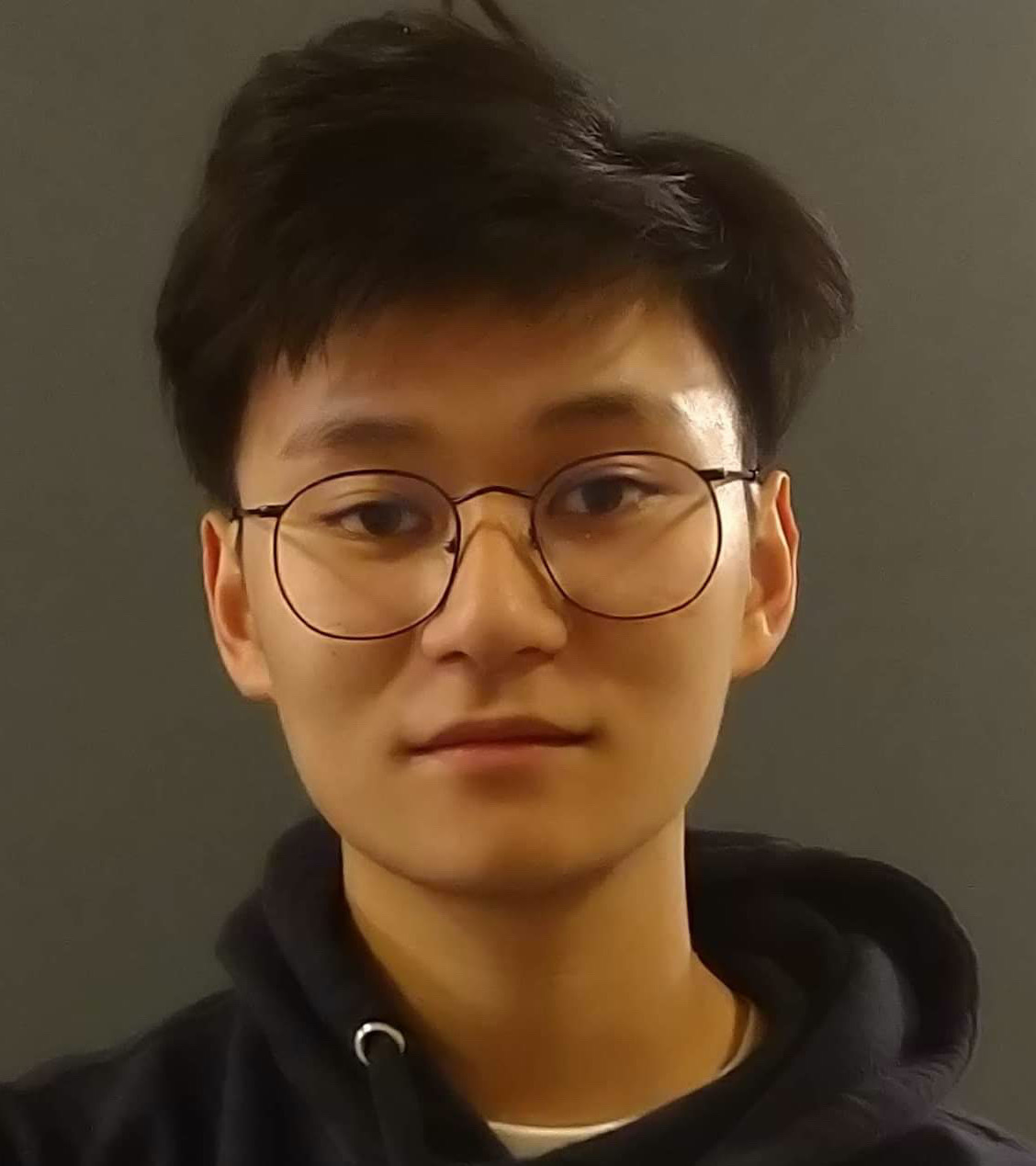}}}]{Victor Liao}
  is an undergraduate student studying Computer Science and Mathematics at the University of Waterloo. He has conducted research at the University of Alberta, the Vector Institute/University of Toronto and the University of Waterloo. His research interests include optimization on Riemannian manifolds, convex and nonconvex optimization, information geometry, and machine learning.
\end{IEEEbiography}

\begin{IEEEbiography}[{\raisebox{0.7cm}{\includegraphics[width=1in,height=1.25in,clip,keepaspectratio]{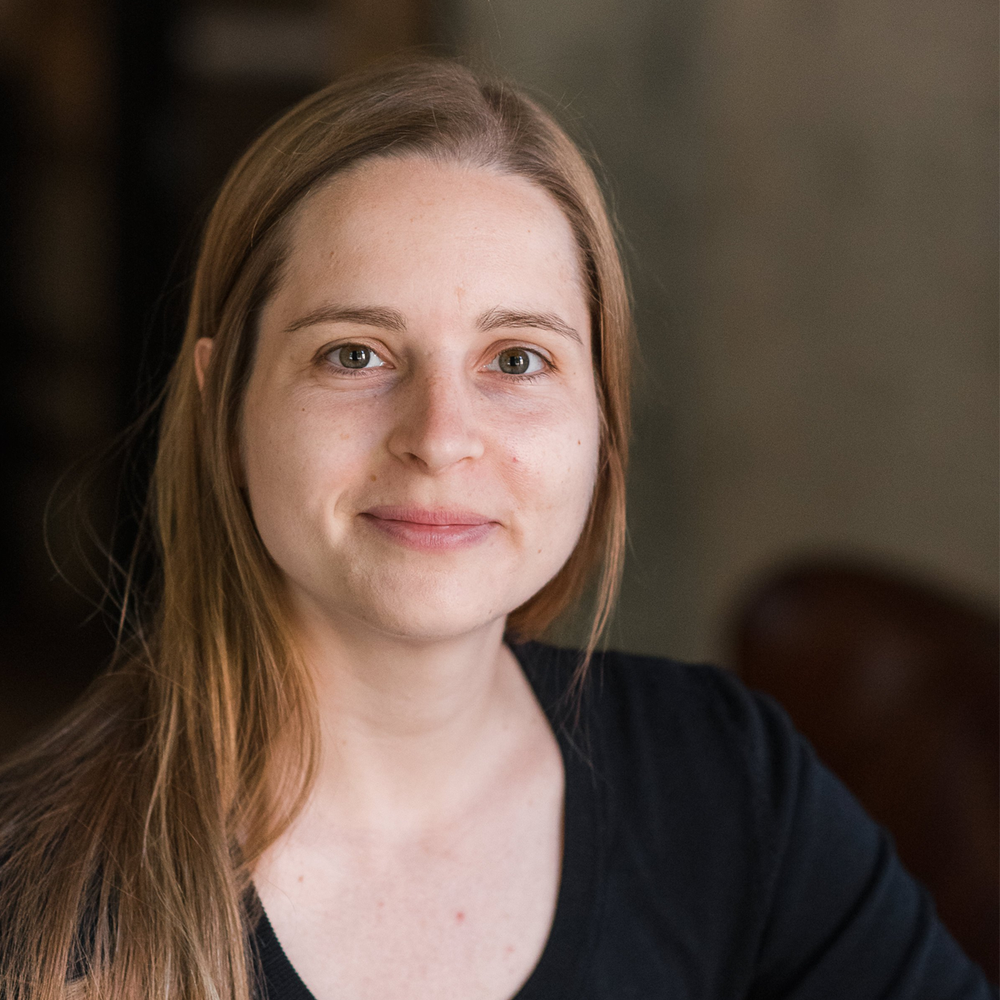}}}]{Martha White}
  is an Associate Professor of Computing Science at the University of Alberta and a PI of Amii (the Alberta Machine Intelligence Institute). She holds a Canada CIFAR AI Chair and received IEEE's ``AIs 10 to Watch: The Future of AI'' award in 2020. She has authored more than 50 papers in top journals and conferences and is an associate editor for TPAMI.
\end{IEEEbiography}

\clearpage

\newpage
\appendices

\section{Biconjugate Forms}\label{app:biconjugates}

\begin{prop}
  The biconjugate of the square function $f(x) = \frac{1}{2} x^2$ is $f^{**}(x) = \max_{h \in \Re} hx - \frac{1}{2}h^2$.
\end{prop}
\begin{proof}
  Recall the definition of the convex conjugate and correspondingly the biconjugate:
  \begin{align*}
    f^*(x) &= \sup_{h \in \Re} \{ hx - f(h) \} \\
    f^{**}(x) &= \sup_{h \in \Re} \{ hx - f^*(h) \}.
  \end{align*}
  Then the conjugate of the square function with dual parameter $a$,
  \begin{align*}
    f^*(x) = \sup_{a \in \Re} xa - \frac{1}{2} a^2.
  \end{align*}
  Applying the convex conjugate again and we obtain
  \begin{align*}
    f^{**}(x) = \sup_{h \in \Re} \left( xh - \sup_{a \in \Re} \left( ha - \frac{1}{2} a^2 \right) \right).
  \end{align*}
  Clearly, the inner supremum is achieved at $a^* = h$, so plugging in the maximizing value of $a$, we obtain
  \begin{align*}
    f^{**}(x) &= \sup_{h \in \Re} \left( xh - h^2 + \frac{1}{2} h^2 \right) \\
    &= \max_{h \in \Re} \left( xh - \frac{1}{2}h^2 \right).
  \end{align*}
  Finally multiplying by two, $2f(x) = x^2$ and $2f^{**}(x) = \max_{h \in \Re} \left( 2xh - h^2 \right)$, arriving at the biconjugate of the square function used in \citet{dai2017learning} and \citet{dai2018sbeed}.
\end{proof}

\begin{prop}
  The biconjugate of the absolute value function $f(x) = |x|$ is $f^{**}(x) = \max_{h \in [-1, 1]} xh$.
\end{prop}
\begin{proof}
  The proof follows the same format as the proof of the biconjugate for the square function.
  Defining the conjugate and biconjugate respectively,
  \begin{align*}
    f^*(x) &= \sup_{a \in \Re} xa - |a|
    = \begin{cases}
      0 & \text{when } |x| \leq 1 \\
      \infty & \text{otherwise}.
    \end{cases} \\
    f^{**}(x) &= \sup_{h \in \Re} xh - f^*(h).
  \end{align*}

  %% Proof that the conjugate equals those cases.
  % First consider the case when $|h| \leq 1$ implying that $h - 1 \leq 0$ and $h + 1 \geq 0$.
  % And when $a \geq 0$, we get
  % \begin{align*}
  %   \max_{a \geq 0} ah - |a| = \max_{a \geq 0} (h - 1) a
  % \end{align*}
  % and because $(h - 1) a \leq 0$, then
  % \begin{align*}
  %   \max_{a \geq 0} (h - 1) a = 0.
  % \end{align*}
  % Likewise when $a \leq 0$, because $h + 1 \geq 0$, we have
  % \begin{align*}
  %   \max_{a \leq 0} ah - |a| = \max_{a \leq 0} (h + 1) a = 0.
  % \end{align*}

  % Now consider the case that $|h| > 1$.
  % When $h > 1$, then $\max_{x \geq 0} ah - |a| = \infty$ since $h + 1 \geq 0$ and
  % when $h < -1$, then likewise $\max_{x \leq 0} ah - |a| = \infty$ since $h - 1 \leq 0$.

  Simplifying the biconjugate form, we get
  \begin{align*}
    f^{**}(x) &= \sup_{h \in \Re} \begin{cases}
      xh & \text{when } |h| \leq 1 \\
      xh - \infty & \text{otherwise}.
    \end{cases} \\
    &= \sup_{|h| \leq 1} xh
    \hspace{.5cm}\anote{$-\infty$ is not feasible}
    \\
    &= \sup_{h \in [-1, 1]} xh.
  \end{align*}
  Finally, considering the maximizing values of $h$, we get $h = -1$ when $x < 0$ and $h = 1$ with $x > 0$ so the biconjugate simplifies to
  \begin{align*}
    f^{**}(x) &= \sign(x)x
    = |x|
    = f(x)
  \end{align*}
  thus completing the proof.
\end{proof}

\subsection{Proof of Theorem~\ref{thm:upper-bound}}\label{app:bound_proof}
We start by showing that the mean absolute Bellman error bounds the mean absolute value error.
We then show in Lemma~\ref{lem:l1-huber-bound} that the Huber function is a close approximation of the absolute function, and so emits an upper-bound with some approximation error controlled by the Huber parameter $\tau$.
Putting these together, then, we show that the mean Huber Bellman error upper bounds the mean absolute \emph{value} error with a small non-vanishing approximation term.

\begin{lem}\label{lem:mabe-mave-bound}
For any vector $v \in \Re^d$, then
\begin{equation*}
  \|v_\pi - v\|_1 \leq \|(I - P_{\pi,\gamma})\|^{-1}_1 \|\Bop v - v\|_1.
\end{equation*}
\end{lem}
\begin{proof}
  First notice that $v_\pi = (I - P_{\pi,\gamma})^{-1} r_\pi$ where $r_\pi$ is the expected reward function with respect to policy $\pi$.
  Then we have
  \begin{align*}
    \Bop v - v  &= r_\pi + P_{\pi,\gamma} v - v && \anote{by definition of $\Bop v$} \\
    &= r_\pi - (I - P_{\pi,\gamma}) v && \anote{rearrange terms to group $v$} \\
    &= (I - P_{\pi,\gamma})v_\pi - (I - P_{\pi,\gamma}) v && \anote{$r_\pi = (I - P_{\pi,\gamma}) v_\pi$} \\
    &= (I - P_{\pi,\gamma}) (v_\pi - v) && \anote{rearrange terms}
  \end{align*}
  Bringing the $(I - P_{\pi,\gamma})$ term to the left side, then
  \begin{align*}
    (I - P_{\pi,\gamma})^{-1} (\Bop v - v) = v_\pi - v.
  \end{align*}
  Finally, because the matrix norm induced by the 1-norm is compatible, we get that
  \begin{align*}
    \|(I - P_{\pi,\gamma})^{-1}\|_1 \|\Bop v - v\|_1 \geq \|v_\pi - v\|_1
  \end{align*}
  thus completing the proof.
\end{proof}

\newcommand{\tcap}{\tau_\text{cap}}

\begin{lem}\label{lem:l1-huber-bound}
  Let $\tcap = \min(\tau, 1)$ and $\tau > 0$, then for any vector $a \in \Re^m$ and $0 < \epsilon \leq \tcap^2$
  \begin{equation*}
    \|a\|_1 \leq \sum_{i=0}^m \frac{\sqrt{\epsilon}}{2\epsilon} \huber{\tau}{a_i} + \frac{\sqrt{\epsilon}}{2}.
  \end{equation*}
\end{lem}
\begin{proof}
  Our goal is to show $|a| \leq C \huber{\tau}{a}$ for any $a \in \Re$.
  First notice that if $|a| \geq \tcap$, then $\huber{\tau}{a} \geq \tcap |a|$ by definition of the Huber function and so we are done with this case.
  However, if $|a| < \tcap$, then $\huber{\tau}{a} = a^2$.
  Should we try to find some constant $C$ such that $|a| \leq C a^2$, then we easily find that $C \geq \frac{1}{|a|}$ which goes to infinity as $a$ goes to zero.
  Instead, we can find $|a| \leq C (a^2 + \epsilon)$ for arbitrary $\epsilon > 0$, which yields $C \geq \frac{|a|}{a^2 + \epsilon}$ which is bounded.
  To find $C$, we have
  \begin{align*}
    C &= \max_{|a| \leq \tcap} \frac{|a|}{a^2 + \epsilon} \\
    &= \max_{0 \leq a \leq \tcap} \frac{a}{a^2 + \epsilon} \\
    &= \frac{\sqrt{\epsilon}}{2\epsilon}
  \end{align*}
  because the maximum is obtained at $a = \sqrt{\epsilon}$.
  We now have that, for $|a| < \tcap$ and $C = \frac{\sqrt{\epsilon}}{2\epsilon}$ then $|a| \leq C \huber{\tau}{a}$.
  Choosing $C \geq 1$ to satisfy the $|a| \geq \tcap$ case, and $C = \frac{\sqrt{\epsilon}}{2\epsilon}$ otherwise, we thus obtain our restriction on $\tcap = \min(\tau, 1)$ completing the proof.
\end{proof}

% \section{Projected Bellman errors}\label{app:projected_errors}

\section{Off-policy Updates}\label{app:off-updates}
For all algorithms, let
\begin{align*}
  \rho_{t} &= \frac{\pi(A_{t} | s_{t})}{b(A_{t} | s_{t})} \\
  \delta_t &= R_{t+1} + \gamma_{t+1} v(S_{t+1}) - v(s_t) \\
  \hweights_{,t+1} &= \hweights_{,t} + \alpha_h \left( \rho_{t} \delta_t - \tilde{h}_{\hweights_{,t}}(s_t) \right) \nabla_{\hweights_{,t}} \tilde{h}_{\hweights_{,t}}(s_t)
\end{align*}
for fixed target policy $\pi$ and behavior policy $b$.

\medskip
\noindent
\textbf{GTD2}
\begin{align*}
  h(s_t) &= \tilde{h}_{\hweights_{,t}}(s_t)\\
  \vweights_{t+1} &= \vweights_{t} + \alpha_v h(s_t) \left( \nabla_{\vweights_{t}} v(s_t) - \rho_t \gamma_{t+1} \nabla_{\vweights_{t}} v(S_{t+1})  \right)
\end{align*}%

\medskip
\noindent
\textbf{GTD2-Huber}
\begin{align*}
  h(s_t) &= \clip{\tau}{\tilde{h}_{\hweights_{,t}}(s_t)}\\
  \vweights_{t+1} &= \vweights_{t} + \alpha_v h(s_t) \left( \nabla_{\vweights_{t}} v(s_t) - \rho_t \gamma_{t+1} \nabla_{\vweights_{t}} v(S_{t+1})  \right)
\end{align*}%

\medskip
\noindent
\textbf{GTD2-Abs}
\begin{align*}
  h(s_t) &= \sign \left(\tilde{h}_{\hweights_{,t}}(s_t) \right)\\
  \vweights_{t+1} &= \vweights_{t} + \alpha_v h(s_t) \left( \nabla_{\vweights_{t}} v(s_t) - \rho_t \gamma_{t+1} \nabla_{\vweights_{t}} v(S_{t+1})  \right)
\end{align*}%

\medskip
\noindent
\textbf{TDC}
\begin{align*}
  h(s_t) &= \tilde{h}_{\hweights_{,t}}(s_t)\\
  \vweights_{t+1} &= \vweights_{t} + \alpha_v \rho_t \left( \delta_t \nabla_{\vweights_{t}} v(s_t) - \gamma_{t+1} h(s_t) \nabla_{\vweights_{t}} v(S_{t+1})  \right)
\end{align*}%

\medskip
\noindent
\textbf{TDC-Huber}
\begin{align*}
  h(s_t) &= \clip{\tau}{\tilde{h}_{\hweights_{,t}}(s_t)}\\
  \vweights_{t+1} &= \vweights_{t} + \alpha_v \rho_t \left( \delta_t \nabla_{\vweights_{t}} v(s_t) - \gamma_{t+1} h(s_t) \nabla_{\vweights_{t}} v(S_{t+1})  \right)
\end{align*}%

\medskip
\noindent
\textbf{TDC-Abs}
\begin{align*}
  h(s_t) &= \sign \left(\tilde{h}_{\hweights_{,t}}(s_t) \right)\\
  \vweights_{t+1} &= \vweights_{t} + \alpha_v \rho_t \left( \delta_t \nabla_{\vweights_{t}} v(s_t) - \gamma_{t+1} h(s_t) \nabla_{\vweights_{t}} v(S_{t+1})  \right)
\end{align*}%

\section{Bias of TDC Gradient}\label{app:tdc_bias}
In this section, we discuss the biased gradient estimate used by gradient-correction methods such as TDC or QRC-Huber.
We show that in the case of linear function approximation, when $h$ is the best linear approximation of $\CE{\delta}{s}$, then the gradient is unbiased for the MSBE.\@
However, this is no longer the case when considering the robust objectives nor is it the case when $h$ must be approximated online.
As has been seen in past results \citep{white2016investigating,ghiassian2020gradient,patterson2022generalized}, this biased gradient does not seem to harm TDC's performance---and in fact the lower variance gradient estimate seems to improve empirical performance---however, our own experiments in Section~\ref{sec:prediction_experiments} suggest that for the MABE the biased gradient estimates can often prevent the gradient-correction algorithm from learning.

Manipulating the gradient of the conjugate Bellman error, we get
\begin{align*}
  &-\nabla_\vweights \delta(\vweights) h(s) = h(s) \nabla_\vweights v(s) - \gamma h(s) \nabla_\vweights v(S') \\
  &= \delta(\vweights)\nabla_\vweights v(s) + \underbrace{(h(s) - \delta(\vweights))\nabla_\vweights v(s)}_{\text{extra term}} - \gamma h(s) \nabla_\vweights v(S').
\end{align*}
Let $h^* = \E{xx\tr}^{-1} \E{x \delta}$ be the optimal linear regression solution for $h$ and let both $h$ and $v$ be parameterized with linear function approximation.
Then because $\nabla_\vweights v(s) = x(s)$, we have
\begin{align*}
  &\CE{(h^* - \delta)\nabla_\vweights(s)}{s} \\
  &\qquad= \CE{x (x\tr h^* - \delta)}{s} \\
  &\qquad= x (x\tr h^* - \CE{\delta}{s}) \\
  &\qquad= xx\tr h^* - x \CE{\delta}{s}
\end{align*}
and in expectation across all states
\begin{align*}
  &\E{(h(S) - \delta)\nabla_\vweights(S)} \\
  &\qquad= \E{xx\tr}h^* - \E{x\delta} \\
  &\qquad= \E{xx\tr}\E{xx\tr}^{-1}\E{x \delta} - \E{x\delta} \\
  &\qquad= \E{x \delta} - \E{x\delta} = 0
\end{align*}
where $x = x(s)$.

Unfortunately, consider the case of the robust objectives.
Instead of the $(h(s) - \delta)$ term, we apply a nonlinear transformation to only $h(s)$.
Intuitively, in the case of the MABE the difference between $\sign(h(s)) - \delta$ can be arbitrarily large.
In the case of the MHBE, the bias due to ignoring this additional term is a function of the clipping parameter $\tau$.
Clearly as $\tau \to \infty$, then $\clip{\tau}{h(s)} \to h(s)$ and the same argument applies as in the MSBE case.

The bias of the gradient estimate used by gradient-correction methods for the MHBE in the case of linear function approximation is
\begin{align*}
  &\E{\| (\clip{\tau}{h^*(s)} - \delta(\vweights)) x \|_\infty} \\
  &\ = \E{|\clip{\tau}{h^*(s)} - \delta(\vweights)| \| x \|_\infty} \\
  &\ \leq \sum_{s \in \States} d(s) \begin{cases}
    0 & \hspace{-1.1cm} \text{when } |h^*(s)| \leq \tau \\
    |\tau\sign(h^*(s)) - \delta(\vweights)| \| x \|_\infty & \text{otherwise}
  \end{cases} \\
  &\ = \sum_{|h^*(s)| > \tau} d(s) |\tau\sign(h^*(s)) - \delta(\vweights)| \| x \|_\infty.
\end{align*}
Because when $|h^*(s)| \leq \tau$, then $|\clip{\tau}{h^*(s)}| = |h^*(s)|$ and we are again in the case of the gradient of the MSBE.
However, when $|h^*(s)| \geq \tau$, then $|\clip{\tau}{h^*(s)}| = \tau$ and we accumulate some bias based on how far $\delta$ is from $\tau$.
When $\E{\delta} = 0$, then likewise $h^*(s) = 0$ because the zero vector is always representable by a linear function approximator (by definition of linearity).
Because $\tau > 0$, then $\clip{\tau}{h^*(s)} = 0$ and the bias is zero, so the fixed point of the algorithm remains unchanged.

\section{Additional Results}\label{app:additional_results}
In this section we include supplementary results to the main body of the paper.
We first investigate the relative ordering of the proposed optimization algorithms when measuring the MAVE instead of the MSVE.
We then motivate empirically why we built our nonlinear control algorithm based on a gradient-correction method instead of a saddlepoint method by investigating the performance of nonlinear GQ on our benchmark control domains.
Finally, we end with two ablation studies investigating the impact of the Huber threshold parameter on each domain  as well as a second ablation investigating the impact of the choice to exclude target networks in the main body of the paper, especially for DQN.

% --> MAVE prediction sensitivities
\begin{figure*}[h!]
  \centering
  \begin{subfigure}[t]{.24\textwidth}
    \includegraphics[width=\textwidth]{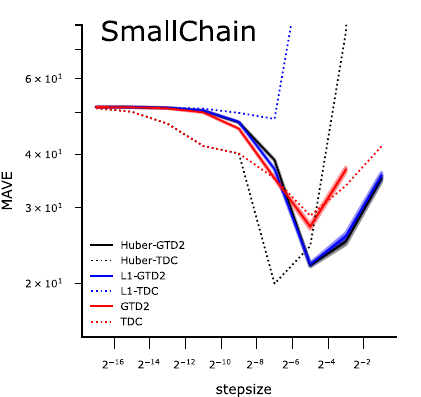}
  \end{subfigure}
  \begin{subfigure}[t]{.24\textwidth}
    \includegraphics[width=\textwidth]{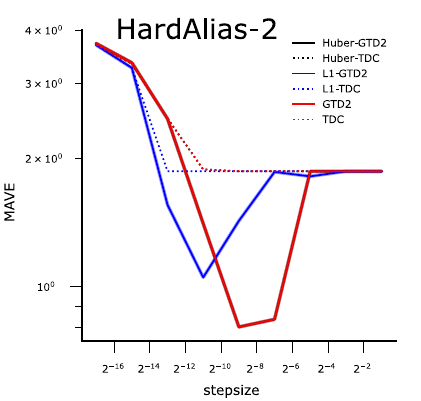}
  \end{subfigure}
  \begin{subfigure}[t]{.24\textwidth}
    \includegraphics[width=\textwidth]{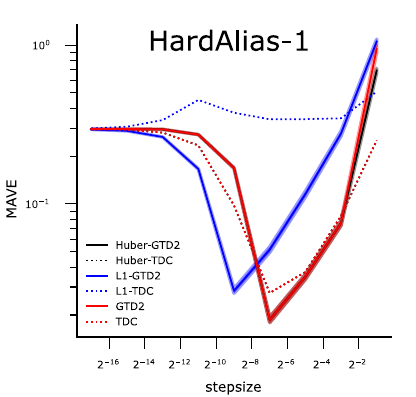}
  \end{subfigure}
  \begin{subfigure}[t]{.24\textwidth}
    \includegraphics[width=\textwidth]{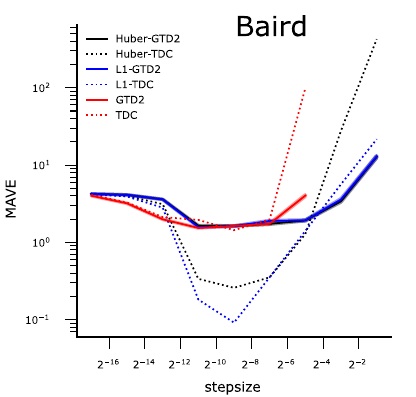}
  \end{subfigure}

  \caption{\label{fig:prediction-MAVE}
    MAVE averaged over 100 independent trials, for each swept stepsize in key prediction domains.
    The mean squared algorithms generally performed well across environments---even the adversarially chosen environments---suggesting the difficulty in minimizing the MABE.
    The Huber algorithms performed best across many environments, often displaying less sensitivity to the choice of stepsize.
  }
\end{figure*}

In Figure~\ref{fig:prediction-MAVE}, we investigate the stepsize sensitivity of each algorithm on four representative linear prediction problems.
The robust algorithms generally have similar best performance as the MSBE algorithms, but with less sensitivity to choice of stepsize.
The $\ell_1$-TDC algorithm often fails to learn a meaningful value function estimate in the given number of training steps.
We hypothesize that this is due to the bias in the gradient estimate for gradient-correction methods, which is pronounced in the case of the MABE but not in the case of the other objectives.
The results in Figure~\ref{fig:prediction-MAVE} are averaged over 100 independent trials for every choice of stepsize, algorithm, and domain.
The shaded regions correspond to standard errors, though error bars are excluded from the TDC algorithms for readability and because the standard errors are negligible.
The performance measure for each algorithm is the average error over the last 25\% of steps in each domain.

% --> Threshold ablation
\begin{figure*}[h!]
  \centering
  \begin{subfigure}[t]{.24\textwidth}
      \includegraphics[width=\textwidth]{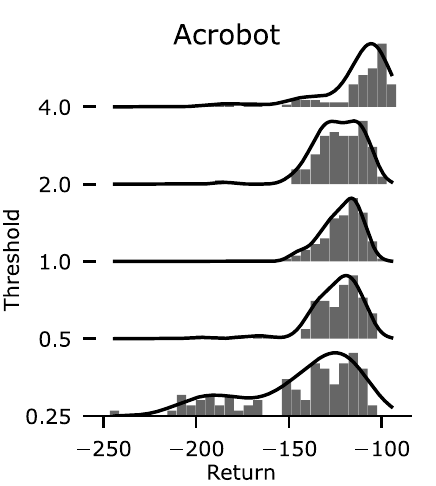}
  \end{subfigure}
  \begin{subfigure}[t]{.24\textwidth}
      \includegraphics[width=\textwidth]{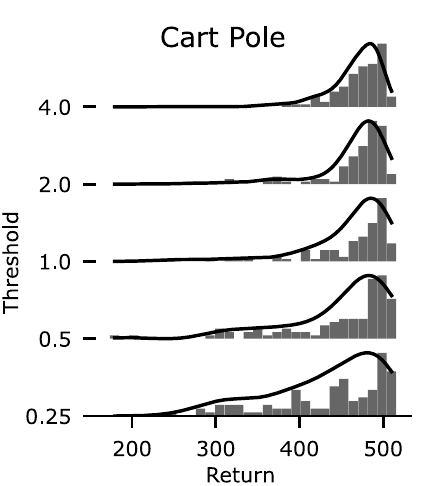}
    \end{subfigure}
  \begin{subfigure}[t]{.24\textwidth}
      \includegraphics[width=\textwidth]{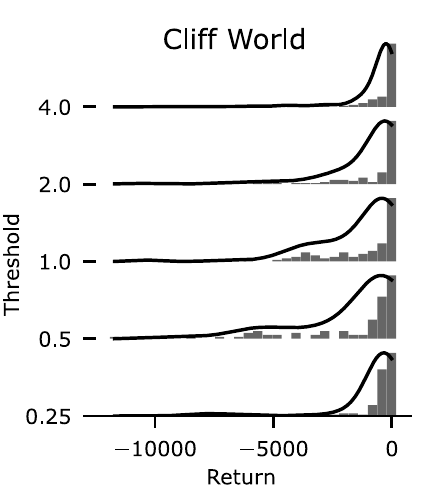}
  \end{subfigure}
  \begin{subfigure}[t]{.24\textwidth}
      \includegraphics[width=\textwidth]{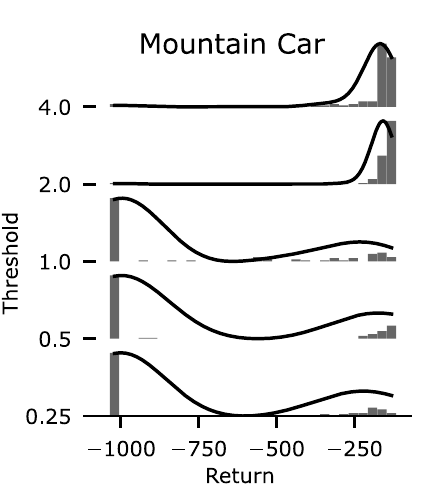}
  \end{subfigure}

  \caption{\label{fig:threshold-ablation}
      Ablating the impact of the threshold parameter for the Huber loss function for the QRC-Huber algorithm across the benchmark domains.
      For three of the domains, QRC-Huber is robust to the choice of threshold parameter with a default value of $\tau = 1$ being a good choice.
      However, the Mountain Car domain shows high-bimodality in performance distribution across multiple random initializations of the neural network for smaller values of the threshold parameter.
  }
\end{figure*}

%% --> Learning curves
%\begin{figure*}
%  \centering
%  \begin{subfigure}[t]{.19\textwidth}
%      \includegraphics[width=\textwidth]{plots/control/learning-curve/Acrobot_auc}
%  \end{subfigure}
%  \begin{subfigure}[t]{.19\textwidth}
%      \includegraphics[width=\textwidth]{plots/control/learning-curve/CartPole_auc}
%    \end{subfigure}
%  \begin{subfigure}[t]{.19\textwidth}
%      \includegraphics[width=\textwidth]{plots/control/learning-curve/CliffWorld_auc}
%  \end{subfigure}
%  \begin{subfigure}[t]{.19\textwidth}
%      \includegraphics[width=\textwidth]{plots/control/learning-curve/MountainCar_auc}
%  \end{subfigure}
%  \begin{subfigure}[t]{.19\textwidth}
%      \includegraphics[width=\textwidth]{plots/control/learning-curve/LunarLander_auc}
%  \end{subfigure}
%
%  \caption{\label{fig:learning-curves}
%    Learning curves for the best meta-parameter for each domain, averaged over 100 independent random trials.
%    Shaded regions indicate one standard error.
%    In Acrobot and Cart Pole, QRC-Huber and QRC have similar performance.
%    In Acrobot and Cliff World, DQN and QRC-Huber have similar performance.
%    However, in Mountain Car and Lunar Lander, QRC-Huber has significantly better performance than both competitors.
%    In total, QRC-Huber is the only algorithm to reliably solve each domain.
%  }
%\end{figure*}

% --> GQ comparison
\begin{figure*}[h!]
  \centering
  \begin{subfigure}[t]{.24\textwidth}
      \includegraphics[width=\textwidth]{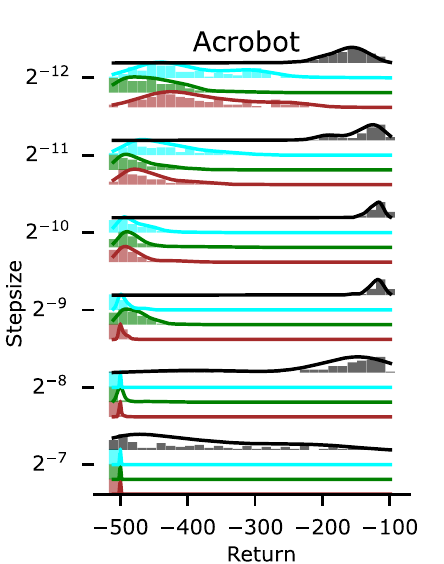}
  \end{subfigure}
  \begin{subfigure}[t]{.24\textwidth}
      \includegraphics[width=\textwidth]{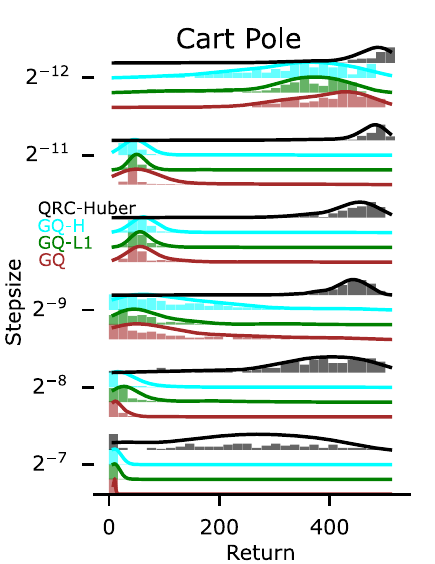}
    \end{subfigure}
  \begin{subfigure}[t]{.24\textwidth}
      \includegraphics[width=\textwidth]{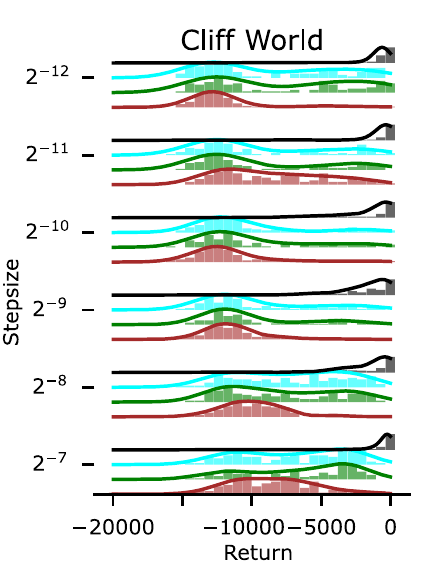}
  \end{subfigure}
  \begin{subfigure}[t]{.24\textwidth}
      \includegraphics[width=\textwidth]{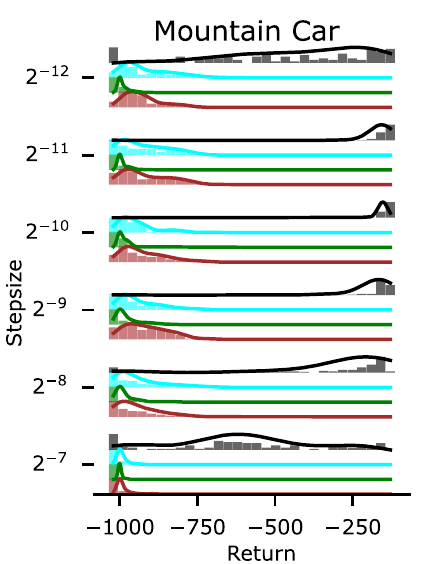}
  \end{subfigure}

  \caption{\label{fig:gq-comparison}
      Comparing the mean return over the last 25\% of steps across several saddlepoint methods against QRC-Huber.
      The saddlepoint methods generally perform very poorly, frequently finding a policy only slightly better than the random policy.
      These results are consistent with the findings of \citet{ghiassian2020gradient} and motivate building on gradient-correction methods for nonlinear control.
      Like QRC-Huber, GQ-Huber uses a twice differentiable estimate of the clip function and all algorithms use the ADAM optimizer.
  }
\end{figure*}

% --> AUC stepsize dists
\begin{figure*}[h!]
  \centering
  \begin{subfigure}[t]{.24\textwidth}
      \includegraphics[width=\textwidth]{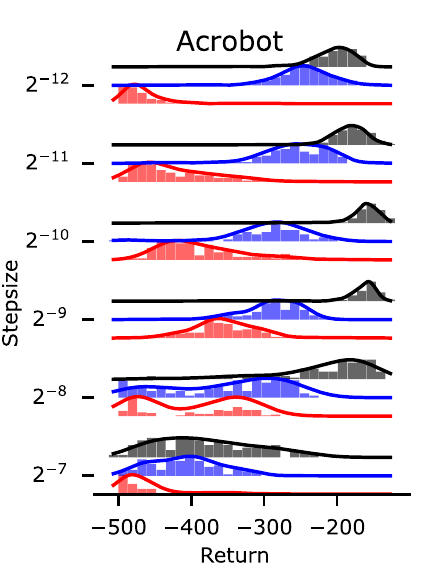}
  \end{subfigure}
  \begin{subfigure}[t]{.24\textwidth}
      \includegraphics[width=\textwidth]{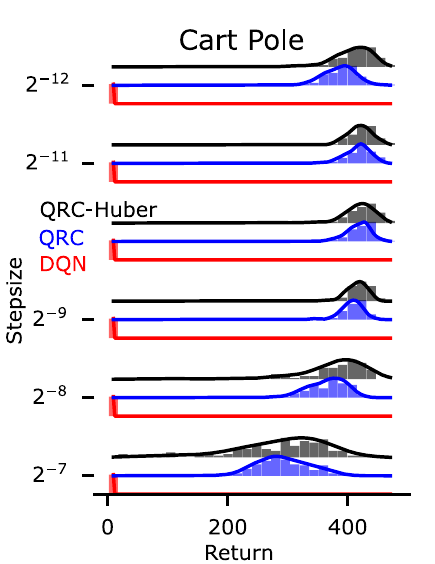}
    \end{subfigure}
  \begin{subfigure}[t]{.24\textwidth}
      \includegraphics[width=\textwidth]{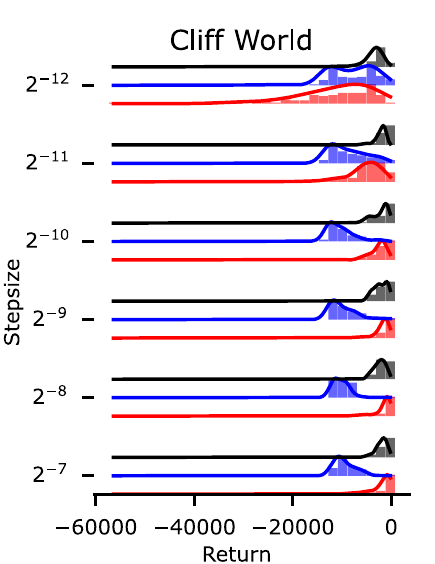}
  \end{subfigure}
  \begin{subfigure}[t]{.24\textwidth}
      \includegraphics[width=\textwidth]{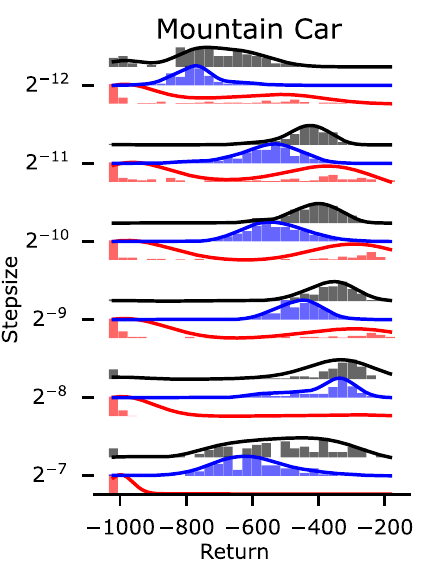}
  \end{subfigure}

  \caption{\label{fig:benchmark-control-auc}
      Comparing algorithms on benchmark control domains with the area under the learning curve as the performance metric.
      Unlike Figure~\ref{fig:benchmark-control}, early learning is included in the performance metric, giving a sense of the sample complexity of each algorithm.
      QRC-Huber tends to perform favorably across all four domains compared to QRC and DQN, exhibiting much more narrow performance distributions that are often centered around higher rewards than the competitor algorithms.
  }
\end{figure*}

% --> Minatar learning curve
\begin{figure}[h!]
  \centering
  \includegraphics[width=0.5\columnwidth]{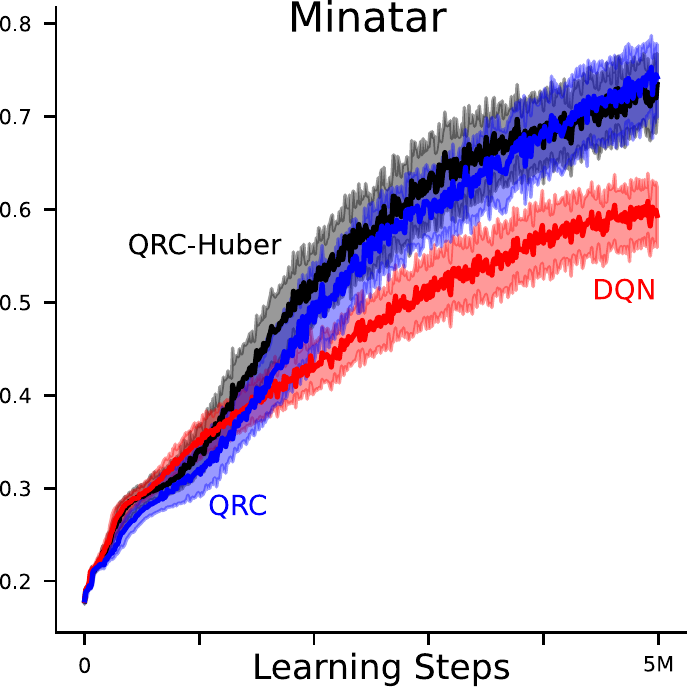}
  \caption{\label{fig:minatar-learning-curve}
    Evaluating the performance of each control algorithm on the Minatar suite of games.
    The learning curves show the scaled performance metric averaged across domains with 95\% bootstrapped confidence intervals about the mean.
    Because each point in the learning curve has less underlying structure than the aggregate performance metric, the confidence intervals are significantly more pessimistic than reported in Table~\ref{tab:minatar}.
    As such, the sample mean performance of QRC-Huber is slightly higher than QRC during early learning, but not statistically significantly so in this result.
    Both gradient-based algorithms considerably outperform DQN with statistical significance, indicating both less domain-sensitivity to meta-parameters as well as better absolute performance.
  }
\end{figure}

Figure~\ref{fig:gq-comparison} investigates the saddlepoint optimization algorithm for nonlinear control across our four benchmark domains.
Generally, the Greedy-GQ algorithm performs considerably worse than gradient-correction algorithms; a motivating factor for building on gradient-correction methods \citep{dai2018sbeed,ghiassian2020gradient}.
A possible explanation for this poor performance is the dependency of the representation learning process on having an accurate estimate of $h(s)$, which itself depends on having a well-learned representation.
This circular dependency is less obviously present in gradient-correction methods, which depend on a sample of the error signal instead of an estimate of the error signal for the primary learning process.
Improving the performance of saddlepoint optimization methods for nonlinear control is important future work.

\subsection{Ablating Design Decisions}\label{app:ablations}

One of the proposed robust objectives depends on a new meta-parameter---$\tau$ the threshold for the Huber function---which was not present in previous extensions of the conjugate Bellman error to nonlinear control.
While for most of our domains we could reasonably pick a default value of $\tau=1$ and avoid allowing our proposed algorithm more opportunities to tune meta-parameters, this choice impacted the claims made in the main body of the paper.
The choice of $\tau$ depends on the magnitude of the TD errors experienced by the algorithm during optimization, which is driven in-part by the magnitude of the rewards, and thus is domain-dependent.
To decrease this dependency, we could consider scaling the magnitude of rewards in a domain-independent way, for instance using PopArt \citep{hessel2018multitask}.

In Figure~\ref{fig:threshold-ablation}, we ablate over several choices of threshold parameter for the QRC-Huber algorithm.
We likewise investigated the sensitivity of DQN to choice of threshold parameter, but due to the general instability of DQN it was less clear which values of $\tau$ were generally best.
Allowing DQN to select its best $\tau$ for each domain does not change the conclusions in Figure~\ref{fig:benchmark-control}, so for simplicity we choose to maintain consistency between DQN and QRC-Huber.
We see long-tail performance distributions as the threshold parameter is made smaller, likely due to the optimization process spending more time in the mean absolute region of the Huber loss.
On the Mountain Car domain, both QRC-Huber and DQN were significantly impacted by $\tau < 2$ and saw strongly bimodal performance distributions.

The poor performance of DQN in Figure~\ref{fig:benchmark-control}---especially on the Cart-pole domain---is surprising; however recent work has also shown that DQN has shockingly poor performance on a wide variety of domains \citep{obando-ceron2021revisiting}.
A potential explanatory factor for the poor performance could be the choice to exclude target networks from our investigation, which could adversely affect DQN disproportionately compared to the gradient-based algorithms.
To understand the impact of our choice to not use target networks, we ablate over the number of steps taken between synchronizing the weights of the target network.

Figure~\ref{fig:target-ablation} shows the number of steps between target network synchronization for QRC-Huber, QRC, and DQN, where one step of synchronization refers to not using target networks at all.
For Acrobot and Cartpole, increasing the number of steps between synchronization appears to harm the performance of the gradient-based methods, likely due to artificially reducing the speed that the bootstrapping targets receive new information.
DQN, on the other hand, benefits from the reduced variance in the bootstrapping targets and tends to perform better across all three domains as the target networks are updated less frequently; though even with 500 steps between synchronization, DQN performs poorly on Mountain Car.

\section{Experimental Details}\label{app:experiment_details}
\subsection{Environments}
In this section we provide further details for the environments and problem settings used in Sections~\ref{sec:prediction_experiments} and~\ref{sec:control_experiments}.

\textbf{HardAlias-1} is an 8-state random walk where the agent starts in the far left state and moves right with 90\% probability.
The episode terminates on taking the move right action in the far right state.
The agent receives $-1$ reward per step with a discount factor of $\gamma=0.99$.
The first, third, and final states all share a common feature, and the remaining five states use the dependent feature representation from \citet{sutton2009fast}, resulting in a feature vector of size $d=4$ for each state.

The second hard aliasing problem, \textbf{HardAlias-2}, is the 2-state problem from \citet{tsitsiklis1997analysis}.
We lightly modify the reward function of the MDP so the optimal value function cannot be perfectly represented, thus allowing each objective to have different minima.
The agent receives a reward of +1 after transitioning from the first state to the second, and a reward of 0 for all other transitions.

\textbf{Outlier} is a large 49-state random walk with an additional ``entry'' state, where the agent has an $\epsilon=0.01$ chance of terminating immediately with -1000, or a $1-\epsilon$ chance of entering the middle state of the random walk.
To emulate a more realistic learning scenario, we use a randomly initialized frozen neural network with ReLU activations and two hidden layers of sizes ten and five respectively to generate five features to describe 50 states.
Taking the left action in the far left state of the random walk results in termination and a reward of $-1$ and correspondingly the right action in the right state results in termination with a reward of $+1$.
The discount factor is set to $\gamma=0.99$ and the left action was chosen with probability $\epsilon$ in every state.

For the two random walks, the first with $N=5$ states (\textbf{SmallChain}) and the other with $N=19$ states (\textbf{BigChain}), we use a randomly initialized neural network representation with ReLU activations and three hidden layers of sizes $h_1 = 4N$, $h_2=N$, and $h_3=\frac{N}{2}$ units respectively, resulting in a feature representation of size $d = h_3$.
These problems are off-policy with the target policy taking the left action with 90\% probability and the behavior policy taking both actions with equal probability.
The discount is $\gamma=0.99$ for both problems.

Finally, \textbf{Baird} is the well-known star MDP from \citet{baird1999reinforcement}. It is used to investigate optimization performance in a high-variance off-policy setting where TD diverges. We do not use it to evaluate the quality of fixed points, because the linear function approximation can represent the true values, and so the fixed-points of each objective are equal in quality.

In the nonlinear experiments, for all domains we use a discount factor of $\gamma=0.99$ and an $\epsilon$-greedy policy with $\epsilon=0.1$.
In Mountain Car, Acrobot, and Cliff World, the agent receives a reward of -1 per step until termination and in Cart-pole the agent receives a reward of +1 per step.
Every environment has an episode cutoff if the agent fails to reach a terminal state with a pre-specified number of steps.
When cut off, the agent is teleported back to the start state and does not update its value function, thus preventing the agent from bootstrapping over the teleportation transition.
All algorithms are run for one hundred thousand steps in total across all episodes, except in Lunar Lander where algorithms are run for 250k steps.

In the Mountain Car environment \citep{moore1990efficient,sutton1996generalization}, the goal is to drive an underpowered car to the top of a hill.
The agent receives as state the position and velocity of the car, and can choose to accelerate forward, backward, or to do nothing on each timestep.
The episode terminates when the agent reaches the top of the hill, or is cut off when the agent reaches a maximum 1000 steps.
In the Cart-pole domain \citep{barto1983neuronlike}, the agent balances a pole attached to a cart which can move along a single axis.
The agent receives as state the position and velocity of the cart, as well as the angle and angular velocity of the pole.
The episode ends when the pole falls or if the agent reaches the maximum of 500 steps.
Finally, the Acrobot domain \citep{sutton1996generalization} has the agent swing a double-jointed arm above a threshold by moving only the inner joint.
The agent receives as state the current angle and velocity of the joints and can take as action, swing left or swing right.
The episode terminates when the specified height is achieved, or is cut off after 500 steps.

The Cliff World environment---lightly adapted from \citet{sutton2018reinforcement}---is a discrete gridworld with 20 states.
The agent starts in the bottom left state and seeks to reach the goal state in the bottom right.
Along the bottom of the grid lies a cliff, where the agent receives a large penalty of -1000 reward for stepping into the pit and is teleported back to the initial state \emph{without} terminating the episode.
The episode terminates only when the agent reaches the goal state, or is cut off when the agent reaches a maximum of 500 steps.

For all environments, we fix meta-parameters other than the stepsize to their default values.
For QRC-Huber, we fix regularizer parameter $\beta=1$ and secondary stepsize ratio $\eta=1$.
For both mean Huber algorithms, we fix the Huber threshold parameter $\tau=1$ for all domains except Mountain Car, where we use $\tau=2$.
We further ablate the impact of this decision in Section~\ref{app:additional_results}.
We sweep over the stepsize parameter for all algorithms and environments and report results for every swept stepsize.

\subsection{Finding Fixed-points}
To find the fixed-points of the objectives in Section~\ref{sec:prediction_experiments}, we used an iterative optimization procedure that assumed access to the underlying dynamics of each MDP to compute exact expected gradients for each update to the primary variable.
We use first order stopping conditions to ensure that the optimization procedure has reached a fixed-point; i.e.\@ when the norm of the gradient is near zero (specifically less than $10^{-7}$).
We use ADAM parameters of $\beta_1 = 0.99$ and $\beta_2 = 0.999$ along with a moving iterate average with exponential moving average parameter $\beta = 0.9$ to reduce oscillation of the gradients and iterates around the fixed-point (especially for the mean absolute objective, where we performed subgradient descent).
We decay the global stepsize according to $\alpha = \frac{1}{\sqrt{t}}$ where $t$ is the number of update steps taken so far.
\footnote{
  We note that the MSBE fixed-point can easily be computed analytically using a least-squares solver.
  For consistency, we use the iterative solver even for the MSBE.\@
  Reported results and conclusions are unchanged when using the analytical solutions.
}

\subsection{Linear Prediction}
For the prediction problems comparing optimization methods in Section~\ref{sec:prediction_experiments}, we swept over the primary and secondary stepsize for all algorithms allowing each to be tuned independently.
We swept values of the primary stepsize $\alpha \in \{2^{-1}, 2^{-2}, \ldots, 2^{-10} \}$ and the ratio between the primary and secondary stepsize $\eta = \frac{\alpha_\theta}{\alpha_h} \in \{2^{-6}, 2^{-4}, \ldots, 2^0, \ldots, 2^6 \}$.
All algorithms have the same number of meta-parameter combinations, so comparison between each algorithm remains fair.
Reported results use SGD with a constant stepsize, though results using RMSProp yielded similar conclusions and thus were omitted.
All algorithms were evaluated after 10k updates for each domain except the random walk which required 100k updates to reasonably converge.

\subsection{Nonlinear Control}
For all of the nonlinear control algorithms and domains, we used neural network function approximation with two hidden layers and ReLU activation units.
For Acrobot, Mountain Car, and Cliff World we used 32 hidden units in both layers, and in Cart Pole we noticed significantly better performance for all algorithms when using 64 hidden units (consistent with the findings of \citet{obando-ceron2021revisiting} which suggested Cart Pole needs considerably more parameters for good performance), finally for Lunar Lander we used 128 hidden units in both layers.
We use experience replay buffers to store the 4000 (10k for Lunar Lander) most recent transitions and draw 32 samples to compute mini-batch updates on every timestep.
We use the ADAM optimizer with default parameters for all algorithms ($\beta_1 = 0.9$ and $\beta_2 = 0.999$), but notice little difference in conclusions when using SGD or RMSProp optimizers.
All agents are trained using $\epsilon$-greedy behavior policies, with $\epsilon=0.1$ for every domain.
Agents are trained for a fixed 100k steps for Acrobot, Mountain Car, and Cart Pole. Cliff World only required 50k steps to achieve good policies, and Lunar Lander required 250k steps.

For the Minatar games, we used the same function approximation architecture and meta-parameter settings as in \citet{young2019minatar}.
Specifically the neural network uses a single convolutional layer with 16 channels, a stride-width of 1, and a kernel-width of 3 followed by a ReLU activation, the output of the convolutional layer is then flattened and sent to a single fully-connected layer with 128 hidden units and ReLU activation.
We use the ADAM optimizer \citep{kingma2015adam} with default parameters and sweep over stepsizes in $\alpha \in \{2^{-13}, 2^{-12}, \ldots, 2^{-8} \}$.
For DQN only, we additionally sweep over target network refresh rates in $\{1, 8, 32, 64\}$ steps.
Experiments are run for 5M steps for each domain and a replay buffer of size 100k is used.

\subsection{Minatar Experimental Procedure}
For the Minatar demonstration, we treat the Minatar domain suite as a single problem setting.
In doing so, we can take advantage of lower variance performance metrics by averaging performance over each of the domains, allowing us to report statistically significant claims using far fewer computational resources.
The procedure is as follows.
\begin{enumerate}
  \item We first swept over several choices of stepsize using only five runs for each game.
  \item We then scaled the AUC for each individual run using probabilistic performance profiles \citep{barreto2010probabilistic} to a value between $[0, 1]$.
  \item We picked the best performing stepsize for each algorithm by averaging the scaled AUC across runs and across games.
  \item Finally we ran an additional 30 runs for each algorithm on each game using that algorithm's best stepsize for a total of 150 runs and report the average scaled performance in Table~\ref{tab:minatar} as well as the average scaled performance over time in Figure~\ref{fig:minatar-learning-curve}.
\end{enumerate}

\subsection{Computational Resources}
For this paper, we used approximately eight CPU years of compute on a general purpose CPU cluster with modern hardware.
We did not use GPUs for any experiment, nor other specialized hardware for training our models.
We used the Torch library \citep{paszke2019pytorch} for defining neural networks and autodifferentiation for the nonlinear control experiments, and used the numpy library \citep{harris2020array} for the linear prediction experiments.

\end{document}